%% file: main.tex
\documentclass[10pt,twocolumn,letterpaper]{article}

\usepackage[pagenumbers]{cvpr} %
\usepackage[accsupp]{axessibility}  %

\input{preamble}

\definecolor{cvprblue}{rgb}{0.21,0.49,0.74}
\usepackage[pagebackref,breaklinks,colorlinks,allcolors=cvprblue]{hyperref}

\def\paperID{9181} %
\def\confName{CVPR}
\def\confYear{2025}

\title{Video Motion Transfer with Diffusion Transformers}

\author{Alexander Pondaven$^1$ \hspace{1em} Aliaksandr Siarohin$^2$ \hspace{1em} Sergey Tulyakov$^2$ \hspace{1em} Philip Torr$^1$ \hspace{1em} Fabio Pizzati$^{1,3}$\\[2mm]
$^1$University of Oxford \hspace{2em} $^2$Snap Inc. \hspace{2em} $^3$MBZUAI\\[2mm]
\url{ditflow.github.io}
}

\begin{document}
\input{sec/teaser}
\input{sec/0_abstract}

\input{sec/1_intro}
\input{sec/2_related}

\input{sec/3_preliminaries}
\input{sec/4_method}

\input{sec/5_experiments}

\input{sec/6_conclusions}
\input{sec/7_acknowledgments}

{
    \small
    \bibliographystyle{ieeenat_fullname}
    \bibliography{main}
}

\input{sec/X_suppl}

\end{document}

%% file: preamble.tex
\usepackage{multirow}
\usepackage{comment}
\usepackage{algorithm}
\usepackage{algorithmic}
\usepackage{makecell}
\usepackage{float}
\usepackage{tikz}
\usepackage[accsupp]{axessibility}
\usepackage{colortbl}

\newcommand{\methodname}{DiTFlow\xspace}
\newcommand{\easy}{Caption\xspace}
\newcommand{\medium}{Subject\xspace}
\newcommand{\hard}{Scene\xspace}

\definecolor{paleRed}{RGB}{250,232,231}
\definecolor{paleBlue}{RGB}{219,226,242}

\usepackage{array}
\newcolumntype{H}{>{\setbox0=\hbox\bgroup}c<{\egroup}@{}}

%% file: sec/teaser.tex
\twocolumn[{
\renewcommand\twocolumn[1][]{#1}
\maketitle
\thispagestyle{empty}
\begin{center}
  \vspace{-0.6cm}
  \centerline{
    \includegraphics[width=\textwidth]{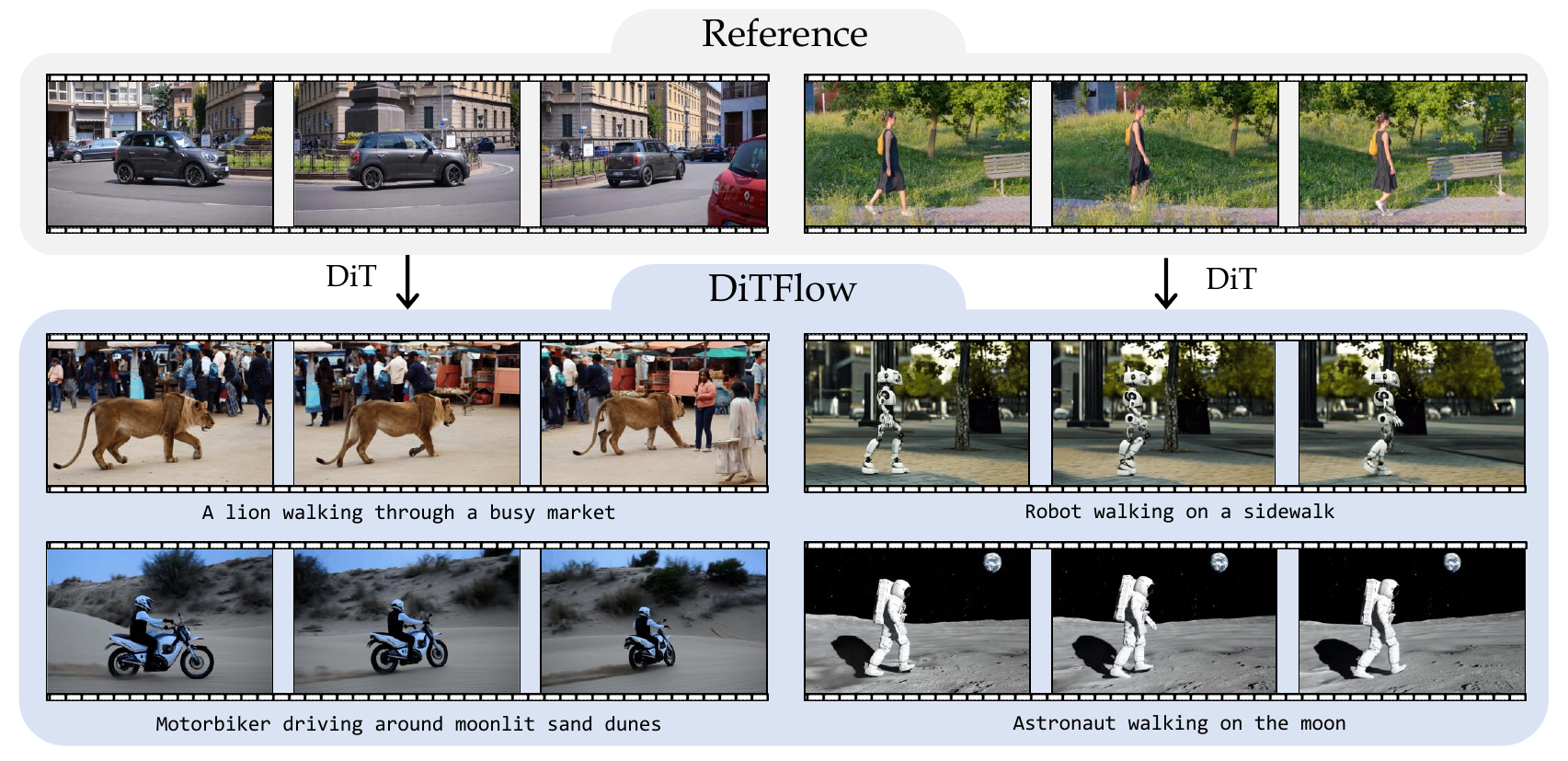}
    \vspace{-0.2cm}
    }
    \captionof{figure}{\textbf{Overview of \methodname.} We propose a motion transfer method tailored for video Diffusion Transformers (DiT). We exploit a training-free strategy to transfer the motion of a reference video (top) to newly synthesized video content with arbitrary prompts (bottom). By optimizing DiT-specific positional embeddings, we can also synthesize new videos in a zero-shot manner.
     }%
    \vspace{-0.1cm}
  \label{fig:teaser}
 \end{center}
}]

%% file: sec/0_abstract.tex
\begin{abstract}

We propose \methodname, a method for transferring the motion of a reference video to a newly synthesized one, designed specifically for Diffusion Transformers (DiT). 
We first process the reference video with a pre-trained DiT to analyze cross-frame attention maps and extract a patch-wise motion signal called the Attention Motion Flow (AMF). We guide the latent denoising process in an optimization-based, training-free, manner by optimizing latents with our AMF loss to generate videos reproducing the motion of the reference one. We also apply our optimization strategy to transformer positional embeddings, granting us a boost in zero-shot motion transfer capabilities. We evaluate \methodname against recently published methods, outperforming all across multiple metrics and human evaluation.\\
\noindent\rule{0.2\textwidth}{0.4pt}\\
\normalfont\scriptsize  Corr.: \texttt{pondaven@robots.ox.ac.uk  fabio.pizzati@mbzuai.ac.ae} %
\end{abstract}

%% file: sec/1_intro.tex
\vspace{-20px}
\section{Introduction}

Diffusion models have rapidly emerged as the global standard for visual content synthesis, largely due to their performance at scale. By scaling the model size, it has been possible to train on increasingly large datasets, even including billions of samples, incredibly boosting synthesis capabilities~\cite{esser2024scalingrectified, videoworldsimulators2024,opensora, yang2024cogvideox}. This trend is especially pronounced in video synthesis, where generating realistic, frame-by-frame visuals with coherent motion relies heavily on extensive data and large models. In this context, a particularly promising development is the introduction of diffusion transformers (DiTs) ~\cite{peebles2023scalable}. Inspired by transformers, DiTs propose a new class of diffusion model that allows for improved scalability, ultimately achieving impressive realism in generation, as demonstrated by their adoption in many scale-oriented open source~\cite{yang2024cogvideox, genmo2024mochi} and commercial systems~\cite{menapace2024snap,polyak2024movie}.

However, realism alone is insufficient for real-world use of synthesized videos. \textit{Control} over the generated video is essential for smooth integration into video creation and editing workflows. Most current models offer text-to-video (T2V) control through prompts, by synthesizing videos aligned with a user's textual description. However, this is rarely sufficient for achieving the desired result. While text may condition the appearance of objects in a scene, it is extremely challenging to control \textit{motion} \ie how the elements move in the scene, since text is inherently ambiguous when describing how fine-grained content evolves over time. To overcome this challenge, \textit{motion transfer} approaches have used existing reference videos as a guide for the dynamics of the scene. The aim is to capture realistic motion patterns and transfer them to synthesized frames. However, most existing approaches are UNet-based~\cite{ronneberger2015unet} and do not take advantage of the superior performance of DiTs, which jointly process spatio-temporal information through their attention mechanism. We believe this opens up opportunities to extract high-quality motion information from the internal mechanics of DiTs.

In this paper, we propose \methodname, the first motion transfer method tailored for DiTs. We leverage the global attention token-based processing of the video, inherent to DiTs, to extract motion patterns directly from the analysis of attention blocks. With this representation, referred to as Attention Motion Flow (AMF), we are able to condition the motion of the synthesized video content, as we show in Figure~\ref{fig:teaser}. We exploit an optimization-based, training-free strategy, coherently with related literature~\cite{yatim2024space,xiao2024video}. In practice, we optimize the latent representation of the video across different denoising steps to minimize the distance to a reference AMF. While employing a separate optimization process for each video yields the best performance, we also discover that optimizing the positional embeddings within DiTs enables the transfer of learned motion to new generations without further optimization, hence in a fully zero-shot scenario not previously possible with UNet-based approaches. This potentially lowers the computational cost of transferring motion on multiple synthesized videos. Overall, our novel contributions are the following:
\begin{enumerate}[topsep=0pt]
    \setlength{\itemsep}{0pt}  %
    \item We propose Attention Motion Flow as guidance for motion transfer on DiTs.
    \item We propose an extension to our optimization objective in this DiT setting, demonstrating zero-shot motion transfer when training positional embeddings.
    \item We test \methodname on state-of-the-art large-scale DiT baselines for T2V, providing extensive comparisons across multiple metrics and user studies.
\end{enumerate}
\label{sec:intro}

%% file: sec/2_related.tex
\section{Related works}
\label{sec:related}

\paragraph{Text-to-video approaches.} Following the success of diffusion models ~\cite{sohldickstein2015, song2020generative, ho2020denoising, song2021score} in generating images from text ~\cite{ramesh2021zeroshot, rombach2022high, ramesh2022hierarchical}, methods to handle the extra temporal dimension in videos were developed ~\cite{ho2022videodiffusionmodels, guo2024animatediff, blattmann2023alignlatents, khachatryan2023zeroscope, blattmann2023svd, chen2024videocrafter2}. These approaches commonly rely on the UNet~\cite{ronneberger2015unet} architecture with separate temporal attention modules operating on solely the temporal dimension for cross-frame consistency. Recently, Diffusion Transformer (DiT) based approaches for text-to-image (T2I) ~\cite{peebles2023scalable, chen2023pixart, esser2024scalingrectified} and text-to-video (T2V) ~\cite{lu2023vdt, gupta2023photo, ma2024latte, chen2024gentron, videoworldsimulators2024,opensora, yang2024cogvideox} have shown superior performance in quality and motion consistency. In particular, VDT \cite{lu2023vdt} highlights the transformer's ability to capture long-range temporal patterns and its scalability. 

\vspace{-8px}
\paragraph{Motion transfer.}
Motion transfer consists of synthesizing novel videos following the motion of a reference one. Unlike video-to-video translation~\cite{saha2024translation, liu2023videop2p, epstein2023diffusionselfguidance}, motion transfer approaches aim for complete disentanglement of the original video structure, focusing on motion alone. Some methods use training to condition on motion signals like trajectories, bounding boxes and motion masks \cite{wang2023videocomposer, yin2023dragnuwa, dai2023animateanything, xing2023makeyourvideo, esser2023structure, wang2024motionctrl, Zhang2024ARXIV}, but this implies significant costs. Other approaches train motion embeddings \cite{Kansy2024ARXIV} or finetune model parameters \cite{zhao2023motiondirector, jeong2023vmc, wu2023tuneavideo, guo2024animatediff}. However, these methods use separate attention for temporal and spatial information, making them unsuitable for DiTs. Optimization-based approaches extract a motion representation at inference \cite{yatim2024space, xiao2024video, hu2024motionmaster, geyer2023tokenflow}, which is more suitable for cross-architecture applications. TokenFlow \cite{geyer2023tokenflow} has a nearest-neighbor based approach on diffusion features, employing expensive sliding window analysis. SMM~\cite{yatim2024space} employ spatial averaging, while MOFT~\cite{xiao2024video} discover motion channels in diffusion features. %

\vspace{-8px}
\paragraph{Attention control in diffusion models.}
Attention features containing semantic correspondences can be manipulated to control generation \cite{hertz2022prompttoprompt, qi2023fatezero, gal2022imageworthwordpersonalizing}. Video editing approaches modify the style or subject with feature injection \cite{bai2024uniedit, ku2024anyv2v, liu2023videop2p, wang2024zeroshotvideoeditingusing} or gradients \cite{motamed2024investigating, epstein2023diffusionselfguidance}. MotionClone~\cite{motionclone} is a UNet-based method that transfers motion by computing a loss on attention. However, this assumes \textit{separate temporal attention} with easily separable motion. This is unavailable for DiTs which use \textit{full spatio-temporal attention} where disentangling motion patterns from content becomes more challenging and requires suboptimal adaptation.

%% file: sec/3_preliminaries.tex
\section{Preliminaries}\label{sec:method-preliminaries}
In this section, we introduce the basic formalism and concepts necessary for \methodname. We begin by reviewing the inference mechanics of T2V diffusion models (Section~\ref{sec:prel-t2v}). We then introduce DiTs for video generation (Section~\ref{sec:prel-dit}).

\subsection{Text-to-video diffusion models}\label{sec:prel-t2v} Let us consider a pre-trained T2V diffusion model. We aim to map sampled Gaussian noise to an output video \( x_0 \) using a denoising network $\epsilon_\theta$ over \( t \in [0, T] \) denoising operations~\cite{ho2020denoising}. To reduce the computational cost of multi-frame generation, video generators typically use Latent Diffusion Models (LDM)~\cite{rombach2022high}, which operate on latent video representations defined by a pre-trained autoencoder~\cite{kingma2022vae} with encoder $\mathcal{E}$ and decoder $\mathcal{D}$. We map a sampled Gaussian $z_T\sim\mathcal{N}(0, I)$ to $z_0 \in \mathbb{R}^{F \times C \times W \times H}$, where $F, C, W, H$ represent the number of frames, latent channels, width, and height, respectively. Noisy latents $z_t$ at each step $t$ maintain the same shape until decoded to the output video $x_0=\mathcal{D}(z_0)$. We formalize the basic denoising iteration as:
\begin{equation}\label{eq:diffusion}
z_\text{t-1} = f(z_t, \epsilon_\theta(z_t, C, t))
\end{equation}
where $C$ is the textual prompt describing the desired output video. This textual signal can condition the network using cross-attentions~\cite{rombach2022high}, or by being directly concatenated with the video latent representation~\cite{yang2024cogvideox}. The function $f$ describes how noise is removed from $z_t$ following a specific noise schedule over T steps~\cite{ho2020denoising,song2020denoising}.\looseness=-1

\subsection{Video generation with DiT}\label{sec:prel-dit}
Unlike U-Net diffusion models~\cite{rombach2022high}, DiT-based systems~\cite{peebles2023scalable} treat the noisy latent as a sequence of tokens, taking inspiration from patching mechanisms typical of Vision Transformers~\cite{dosovitskiy2020image}. The denoising network $\epsilon_\theta$ is replaced with a transformer-based architecture. Latent patches of size $P\times P$ are encoded into tokens with dimension $D$ and reorganized into a sequence of shape $(F\cdot\frac{H}{P}\cdot\frac{W}{P})\times D$. The denoising network $\epsilon_\theta$ is composed of $N$ DiT blocks~\cite{peebles2023scalable} consisting of multi-head self-attention~\cite{transformer} and linear layers. To encode positional information between patches during attention, a positional embedding $\rho$, consisting of values dependent on the patch location in the sequence, conditions the denoising network $\epsilon_\theta(z_t, C, t, \rho)$. Various position encoding schemes exist ~\cite{transformer, shaw2018relative, rope} where $\rho$ is most commonly either added directly to patches at the start of $\epsilon_\theta$ or augment the queries and keys at each attention block.\looseness=-1

%% file: sec/4_method.tex
\begin{figure}[t]
    \centering
    \includegraphics[width=\linewidth, ]{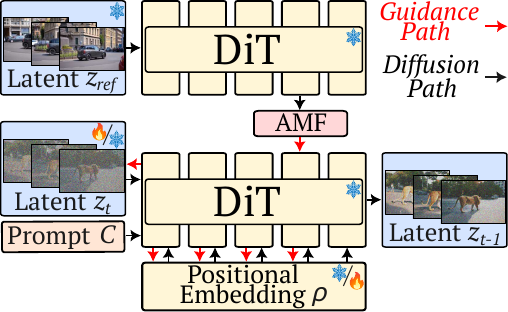}
    \caption{\textbf{Core idea of \methodname.} We extract the AMF from a reference video and we use that to guide the latent representation $z_t$ towards the motion of the reference video. In our experiments, we also tested optimizing positional embeddings for improved zero-shot performance.}
    \label{fig:inference}
\end{figure}

\section{Method}\label{sec:method}

Our core idea is to take advantage of the attention mechanism in DiTs to extract motion patterns across video frames in a zero-shot manner. Building on the intuition behind motion cue extraction discussed in Section~\ref{sec:method-attentions}, we then describe the creation of AMFs in Section~\ref{sec:method-motionflow}. The AMF extracted from a reference video can be used as an optimization objective at a particular transformer block in the denoising network, as illustrated in Figure~\ref{fig:inference}. We define how we use the AMF for optimization in Section~\ref{sec:method-training}. Note that the extracted motion patterns are independent from the input content, enabling the application of motion from a reference video to arbitrary target conditions.

\subsection{Cross-frame Attention Extraction}\label{sec:method-attentions}

We aim to extract the diffusion features of a reference video $x_\text{ref}$ with a pretrained T2V DiT model in order to obtain a signal for motion. The motion of subjects in a video may be described by highly correlated content that changes spatially over time, so \textit{tokens with similar spatial content will intuitively attend to each other across frames}. Hence, we can benefit from extracted token dependencies between frames to reconstruct how a specific element will move over time. We start by computing the latent $z_\text{ref}=\mathcal{E}(x_\text{ref})$ and pass it through the transformer at denoising step $t=0$ with an empty textual prompt. Previous work~\cite{xiao2024video} have observed a cleaner motion signal at lower denoising steps and we found $t=0$ to be suitable for feature analysis of the input video. This also avoids the need for expensive DDIM inversion~\cite{song2020denoising} for our task. Let us consider the $n\text{-th}$ DiT block of $\epsilon_\theta$. The self-attention layer computes keys and queries for $M$ attention heads. We average over the heads for feature extraction to reduce noise and memory consumption when optimizing. We denote $K$ and $Q$ as the keys and queries, averaged across heads, with shape $(F \cdot S, D_h)$, where $S = \frac{H}{P}\cdot\frac{W}{P}$ represents the spatial token length and $D_h = D / M$. We represent this operation as:
\begin{equation}
    \{Q,K\} \xleftarrow{\text{n}} \epsilon_\theta(z_\text{ref}, \varnothing, 0, \rho)
\end{equation} 
Hence, for two frames $(i,j), i,j\in[1, F]$ we can calculate the \textit{cross-frame attention} $A^\otimes_{i, j}$ as follows:
\begin{equation}\label{eq:cross-attention}
    A^\otimes_{i, j} = \sigma \left( \tau \frac{Q_iK_j^T}{\sqrt{d_k}} \right) \in \mathbb{R}^{S \times S}
\end{equation}
In Equation~\eqref{eq:cross-attention}, $Q_i$ and $K_j$ refer to the query and key matrices of the $i\text{-th}$ and $j\text{-th}$ frames with shape $S \times D_h$. Here, $\sigma$ is the softmax operation over the final dimension \ie over tokens in the $j\text{-th}$ frame, $\sqrt{d_k}$ is the attention scaling term~\cite{transformer} and $\tau$ is a temperature parameter. Intuitively, $A^\otimes_{i,j}$ encodes the relationship between patches of frames $i$ and $j$ from $x_\text{ref}$, serving as a signal to capture the reference video motion.\looseness=-1

\begin{figure}[t]
    \centering
    \includegraphics[width=\linewidth]{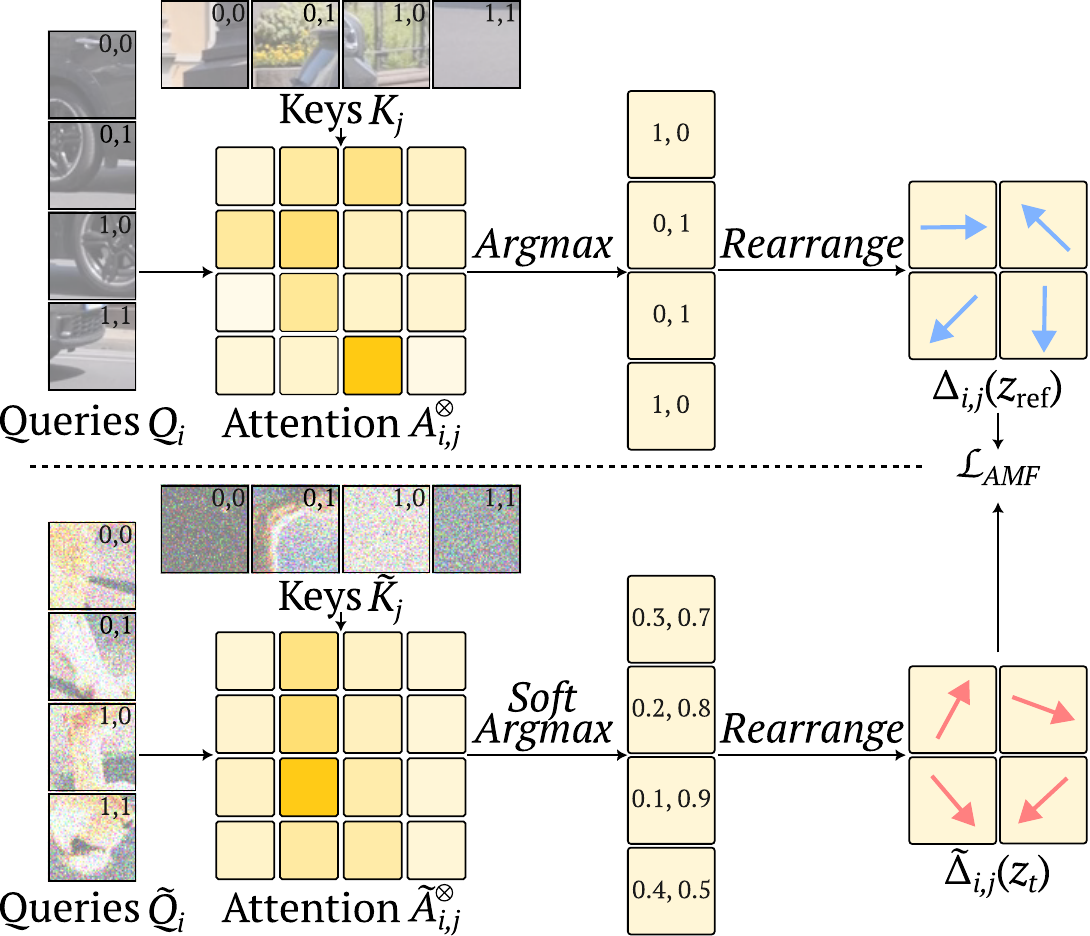}
    \caption{\textbf{Guidance.} We compute the reference displacement by processing cross-frame attentions with an argmax operation and rearranging them into displacement maps, identifying patch-aware cross-frame relationships. For video synthesis, we do the same operation with a soft argmax to preserve gradients, and impose reconstruction with the reference displacement.\looseness=-1}
    \label{fig:displacement}
\end{figure}

\subsection{Attention Motion Flows}\label{sec:method-motionflow}

We subsequently capture the AMF of a video by estimating a displacement map of spatial patches across all frame combinations. Each frame is composed of $\frac{H}{P} \times \frac{W}{P}$ patches. Our goal is to understand how the content of all patches in frame $i$ moves to obtain frame $j$. To achieve this, we first process $A^\otimes_{i,j}$ using an argmax operation, which assigns each patch in frame $i$ to the index of the most attended patch in frame $j$. We denote this result as $\hat{A}^\otimes_{i,j}$ where each entry $\hat{A}^\otimes_{i,j}[(u,v)]$ stores the assigned coordinate $(u', v')$. Empirically, selecting a single index using argmax results in a cleaner displacement map, leading to more reliable motion guidance. Using the obtained index pairs, we construct a patch displacement matrix $\Delta_{i,j}$ of size $S \times 2$ where $\Delta_{i,j}[(u,v)] = (u'-u, v'-v)$. This process is illustrated in Figure~\ref{fig:displacement} (top). Finally, we aggregate the displacement matrices for all frame pairs $(i, j)$ to construct the reference AMF, which serves as the motion guidance signal:  

\begin{equation}
    \text{AMF}(z_\text{ref}) = \{\Delta_{i,j} \mid i,j \in [1, F] \}
\end{equation}

\noindent Our extracted AMF follows the idea of motion vectors used in MPEG-4 patch-based video compression~\cite{richardson2004h}, but is applied to DiT latent representations. In summary, for each patch, we calculate a motion vector in a two-dimensional coordinate space that indicates where the patch will move from frame $i$ to frame~$j$, effectively capturing the motion.

\begin{algorithm}[t]
\caption{\methodname inference pipeline}
\label{alg:motion-transfer}
\begin{algorithmic}[1]
\REQUIRE Reference video $x_\text{ref}$, trained DiT model $\epsilon_\theta$, encoder $\mathcal{E}$, decoder $\mathcal{D}$, prompt $C$, positional embedding $\rho$.
\ENSURE Generated video $x_0$ with transferred motion
\STATE Extract latent representation: $z_\text{ref} \gets \mathcal{E}(x_\text{ref})$
\STATE Compute attention: $\{Q, K\} \xleftarrow{n} \epsilon_\theta(z_\text{ref}, \varnothing, 0, \rho)$
\FOR{each $(i, j)$ where $i, j \in [1, F]$}
    \STATE Calculate cross-frame attention $A^\otimes_{i, j}$ 
    \STATE Construct displacement matrix $\Delta_{i, j}$
\ENDFOR
\STATE Construct reference AMF: $\text{AMF}(z_\text{ref}) \gets \Delta_{i, j}$
\STATE Initialize $z_T \sim \mathcal{N}(0, I)$
\STATE Initialize $\rho_T = \rho$
\FOR{denoising step $t = T$ to $0$}
    \IF{$t > T_\text{opt}$}
        \FOR{optimization step $k = 0$ to $K_\text{opt}$}
            \STATE Extract $\tilde{Q}$ and $\tilde{K}$: $\{\tilde{Q}, \tilde{K}\} \xleftarrow{n} \epsilon_\theta(z_t, C, t, \rho_t)$
            \FOR{each $(i, j)$ where $i, j \in [1, F]$}
            \STATE Calculate cross-frame attention $\tilde{A}^\otimes_{i, j}$
            \STATE Compute soft displacement matrix $\tilde{\Delta}_{i, j}$
            \ENDFOR
            \STATE Construct $\text{AMF}(z_t) \gets \tilde{\Delta}_{i, j}$
            \STATE Get $\mathcal{L}_\text{AMF} \gets ||\text{AMF}(z_\text{ref}) - \text{AMF}(z_t)||^2_2$
            \STATE Update $z_t$ or $\rho_t$ by minimizing $\mathcal{L}_\text{AMF}$
        \ENDFOR
    \ELSE
        \STATE $\rho_t = \rho$
    \ENDIF
    \STATE $z_{t-1} = f(z_t, \epsilon_\theta(z_t, C, t, \rho_t))$
    \ENDFOR
 
\RETURN $x_0 = \mathcal{D}(z_0)$
\end{algorithmic}
\end{algorithm}

\subsection{Guidance and Optimization}\label{sec:method-training} 
Once the reference AMF is obtained from $x_\text{ref}$, we use it to guide the generation of new video content with a T2V DiT model. Specifically, we aim to guide the denoising of $z_T\sim\mathcal{N}(0, I)$ in such a way that $x_0 =\mathcal{D}(z_0)$ reproduces the same motion patterns as $x_\text{ref}$. Here, we enforce guidance with an optimization-based strategy, aimed to reproduce the same extracted AMF at a given transformer block for the generated video in intermediate denoising steps $t$. For a given $t$, we consider key $\tilde{K}$ and query $\tilde{Q}$ extracted by the $n\text{-th}$ DiT block while processing the input latent~$z_t$:
\begin{equation}
    \{\tilde{Q}, \tilde{K}\} \xleftarrow{n}\epsilon_\theta(z_t, C, t, \rho_t)
\end{equation}
We follow the procedure described in Equation~\eqref{eq:cross-attention} to extract the corresponding cross-frame attention $\tilde{A}^\otimes_{i,j}$:
\begin{equation}
\tilde{A}^\otimes_{i,j} = \sigma\left(\tau \frac{\tilde{Q}_i\tilde{K}_j^T}{\sqrt{d_k}}\right)
\end{equation}
We then calculate \textit{soft} displacement matrices $\tilde{\Delta}_{i,j}$ as illustrated in Figure~\ref{fig:displacement} (bottom). Rather than using argmax, we perform a weighted sum of attention values to identify continuous displacement values that preserve gradients where:

\begin{equation}
\tilde{\Delta}_{i,j}[(u,v)] = \sum_{(u', v')} \tilde{A}^\otimes_{i,j}[(u,v), (u', v')] \cdot (u' - u, v' - v)
\end{equation}

\noindent We then build the soft AMF for the current step $t$ as $\text{AMF}(z_t) = \{\tilde{\Delta}_{i,j} \mid i,j \in [1, F] \}$. Finally, we minimize the element-wise Euclidean distance between the AMF displacement vectors of the reference and current denoising step. This equates to minimizing the following loss:
\begin{equation}
    \mathcal{L}_\text{AMF}(z_\text{ref}, z_t) = ||\text{AMF}(z_\text{ref}) - \text{AMF}(z_t)||^2_2
\end{equation}
We follow previous U-Net-based approaches~\cite{xiao2024video,yatim2024space} and optimize $z_t$ by backpropagating this loss in specific denoising steps. An alternative approach, benefiting from the DiT-specific presence of positional encodings, is to backpropagate the loss to optimize positional embeddings $\rho_t$. Note that positional embeddings are responsible for encoding the spatiotemporal locations of content. Intuitively, by manipulating the positional information of latent patches, we \textit{guide the reorganization of patches} for motion transfer. This is also disentangled from latents that encode content. Hence, we empirically observed that while latent optimization leads to better overall performance, optimizing $\rho_t$ enables better generalization of the learned embeddings to new prompts without repeating the optimization, allowing for fully zero-shot inference. In practice, we optimize $z_t$ or $\rho_t$ for $K_\text{opt}$ steps, up until a given denoising step $T_\text{opt}\in[0, T)$ of the diffusion process, while denoising normally for the remaining steps. We report our full inference scheme in Algorithm~\ref{alg:motion-transfer}.

%% file: sec/5_experiments.tex
\begin{table*}[ht]
    \centering
    \setlength{\tabcolsep}{3px} %

    \resizebox{\linewidth}{!}{
    \begin{tabular}{l||cc|cc|cc|cc||cc|cc|cc|cc}
        \multicolumn{1}{c}{} & \multicolumn{8}{c}{\textit{\large{CogVideoX-5B}}}& \multicolumn{8}{c}{\textit{\large{CogVideoX-2B}}} \\    \toprule
         \multirow{2}{*}{\textbf{Method}} & \multicolumn{2}{c|}{\textbf{\easy}} & \multicolumn{2}{c|}{\textbf{\medium}} & \multicolumn{2}{c|}{\textbf{\hard}} &
         \multicolumn{2}{c||}{\textbf{All}} & \multicolumn{2}{c|}{\textbf{\easy}} & \multicolumn{2}{c|}{\textbf{\medium}} &
         \multicolumn{2}{c|}{\textbf{\hard}} & \multicolumn{2}{c}{\textbf{All}} \\
         & MF $\uparrow$ & IQ $\uparrow$ & MF $\uparrow$ & IQ $\uparrow$ & MF $\uparrow$ & IQ $\uparrow$ &
           MF $\uparrow$ & IQ $\uparrow$ & MF $\uparrow$ & IQ $\uparrow$ & MF $\uparrow$ & IQ $\uparrow$ &
           MF $\uparrow$ & IQ $\uparrow$ & MF $\uparrow$ & IQ $\uparrow$ \\\midrule

         Backbone & 0.524 & \underline{0.315} & 0.502 & \textbf{0.321} & 0.544 & 0.318 &
                       0.523 & \underline{0.318} & 0.521 & \underline{0.313} & 0.495 & 0.312 &
                       0.523 & 0.314 & 0.513 & 0.313 \\
         \midrule
         Injection~\cite{xiao2024video} & 0.608 & \underline{0.315} & 0.581 & \textbf{0.321} & 0.635 & \textbf{0.320} &
           0.608 & \textbf{0.319} & 0.546 & \textbf{0.315} & 0.524 & \underline{0.317} &
           0.563 & \textbf{0.321} & 0.544 & \textbf{0.318} \\
        MotionClone~\cite{motionclone} & 0.635 & 0.313 & 0.640 & \textbf{0.321} & 0.628 & \textbf{0.320} &
       0.634 & \underline{0.318} & 0.564 & 0.303 & 0.557 & 0.304 &
       0.574 & 0.301 & 0.565 & 0.303 \\
         SMM~\cite{yatim2024space} & \underline{0.782} & 0.313 & \underline{0.741} & \underline{0.317} & \underline{0.776} & 0.316 &
                                    \underline{0.766} & 0.315 & \textbf{0.687} & 0.312 & \underline{0.682} & 0.312 &
                                    \underline{0.694} & \underline{0.317} & \underline{0.688} & 0.312 \\
         MOFT~\cite{xiao2024video} & 0.728 & 0.313 & 0.728 & \textbf{0.321} & 0.722 & \underline{0.319} &
                                    0.726 & \underline{0.318} & 0.503 & 0.312 & 0.502 & 0.313 &
                                    0.508 & 0.315 & 0.504 & 0.312 \\
        
         \rowcolor{gray!20}\methodname & \textbf{0.790} & \textbf{0.316} & \textbf{0.775} & \textbf{0.321} & \textbf{0.789} & \underline{0.319} &
                      \textbf{0.785} & \textbf{0.319} & \underline{0.685} & 0.311 & \textbf{0.753} & \textbf{0.322} &
                      \textbf{0.739} & \underline{0.320} & \textbf{0.726} & \underline{0.317} \\\bottomrule

    \end{tabular}             
    }    
    \caption{\textbf{Metrics evaluation.} We compare \methodname across 3 different caption setups (\easy, \medium, \hard) and against 4 baselines. We consistently score first or second in all metrics for almost all scenarios, advocating the quality of our motion transfer. Performance is consistent across two backbones with 5B and 2B parameters respectively. Best results are in \textbf{bold} and second best are \underline{underlined}.} 
    \label{tab:quant}
\end{table*}

\begin{figure*}[t]
    \centering
    \setlength{\tabcolsep}{2px}
    \resizebox{\linewidth}{!}{

    \begin{tabular}{ccHc@{\hspace{20px}}cHc@{\hspace{20px}}cHc} %
        &\multicolumn{3}{c}{\hspace{-25px}\textbf{\easy}} &         \multicolumn{3}{c}{\hspace{-25px}\textbf{\medium}} &         \multicolumn{3}{c}{\textbf{\hard}} \\
        \multirow{1}{*}[34px]{\rotatebox{90}{Reference}} & 
        \includegraphics[width=100px,height=50px]{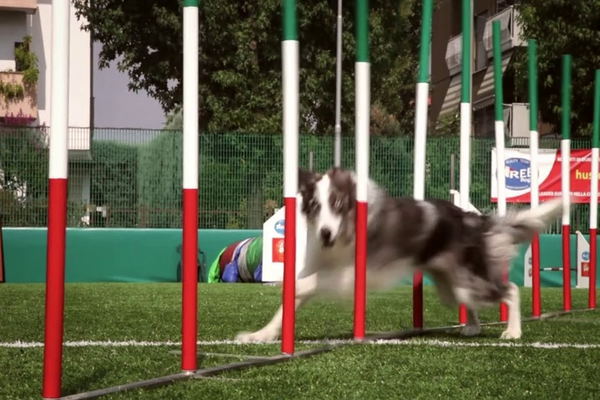} &
        \includegraphics[width=100px,height=50px]{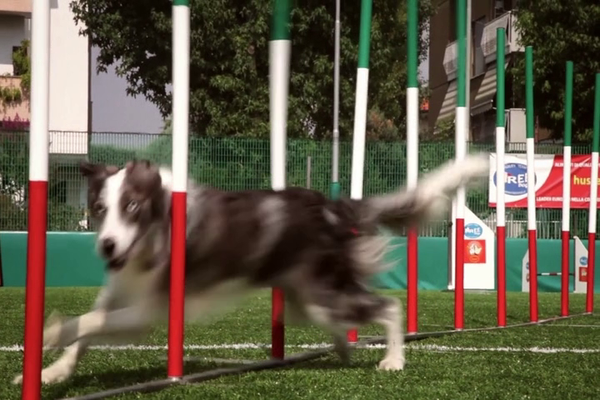} &
        \includegraphics[width=100px,height=50px]{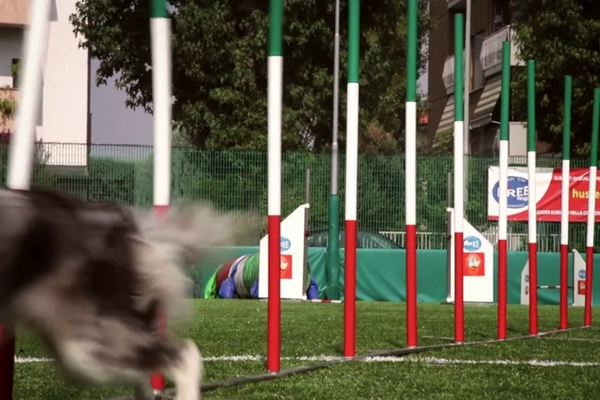} &
        \includegraphics[width=100px,height=50px]{figures/libby_medium/ref1.png} &
        \includegraphics[width=100px,height=50px]{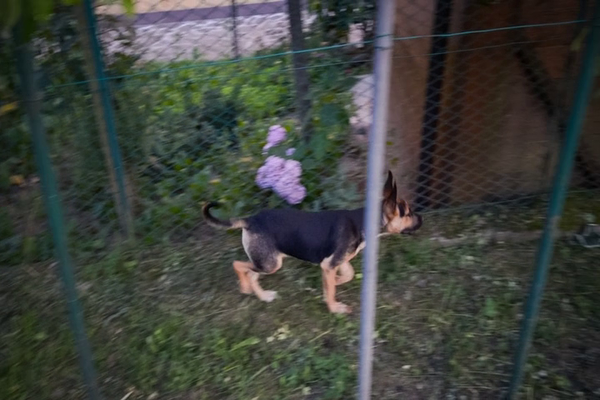} &
        \includegraphics[width=100px,height=50px]{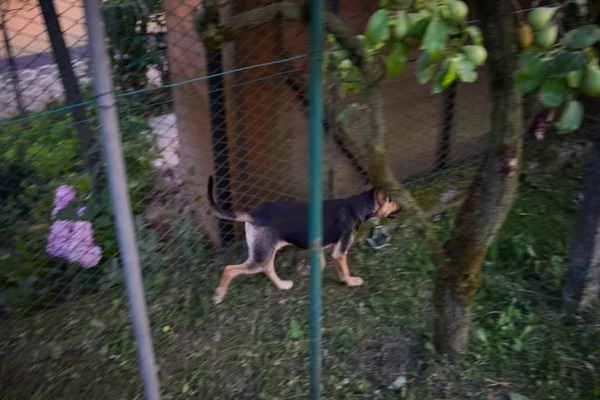} &
        \includegraphics[width=100px,height=50px]{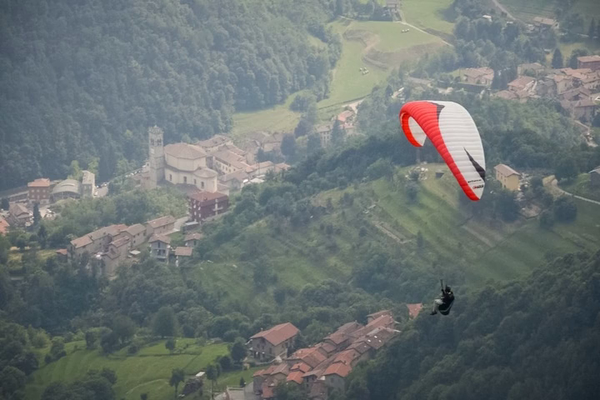} &
        \includegraphics[width=100px,height=50px]{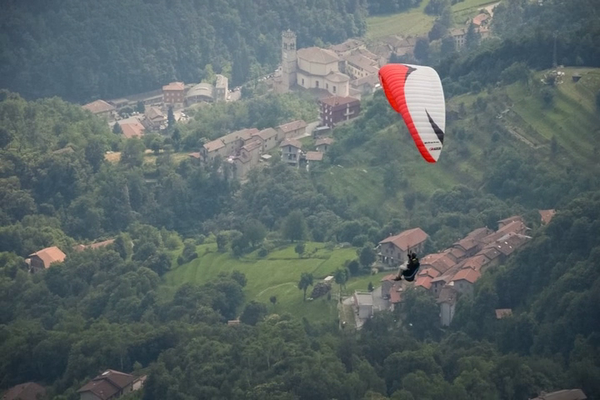} &
        \includegraphics[width=100px,height=50px]{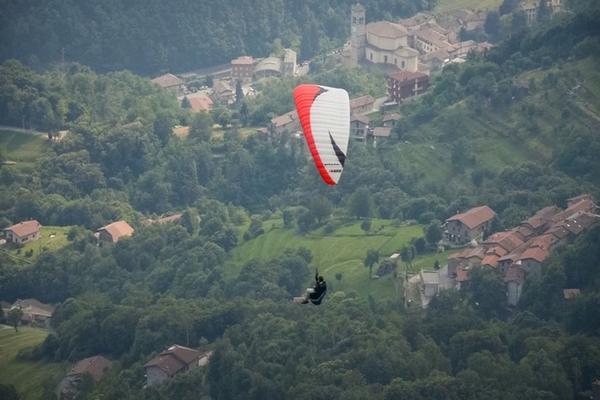} \\ %
        \midrule
        \multirow{1}{*}[40px]{\rotatebox{90}{MClone~\cite{motionclone}}} & 
        \includegraphics[width=100px,height=50px]{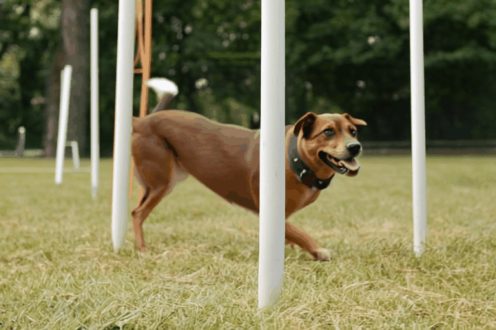} &
        \includegraphics[width=100px,height=50px]{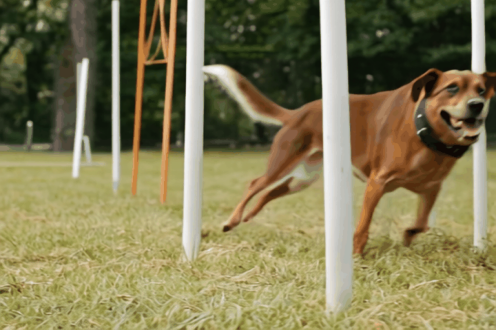} &
        \includegraphics[width=100px,height=50px]{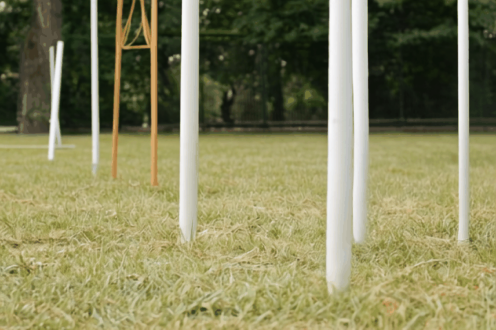} &
        \includegraphics[width=100px,height=50px]{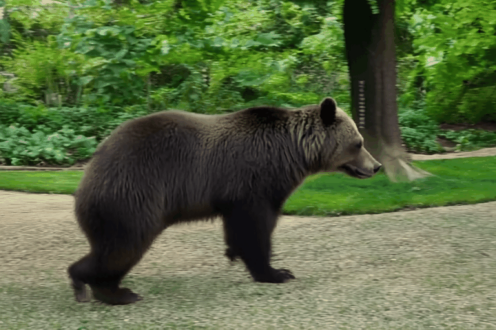} &
        \includegraphics[width=100px,height=50px]{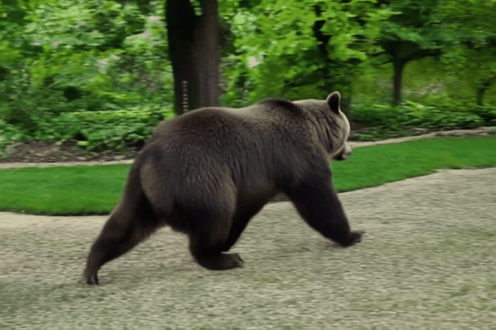} &
        \includegraphics[width=100px,height=50px]{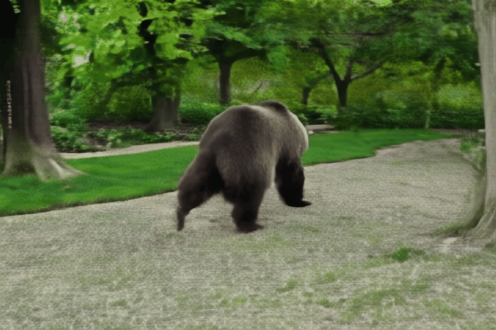} &
        \includegraphics[width=100px,height=50px]{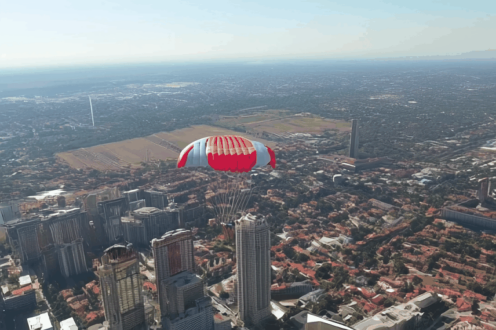} &
        \includegraphics[width=100px,height=50px]{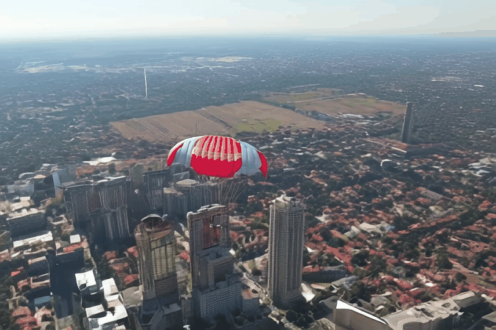} &
        \includegraphics[width=100px,height=50px]{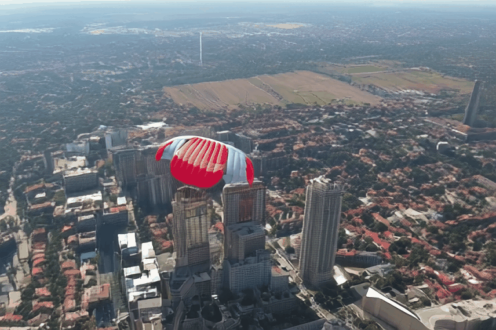} \\ %
        \multirow{1}{*}[34px]{\rotatebox{90}{SMM~\cite{yatim2024space}}} & 
        \includegraphics[width=100px,height=50px]{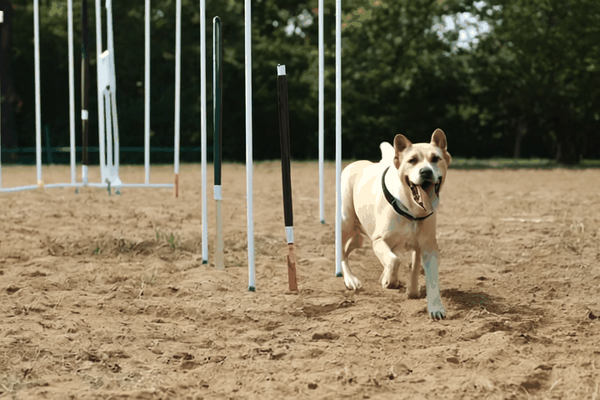} &
        \includegraphics[width=100px,height=50px]{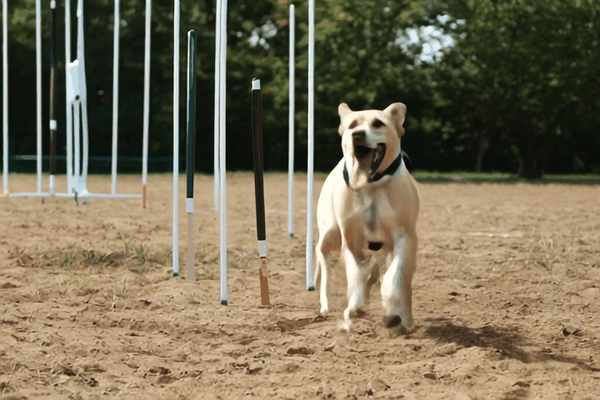} &
        \includegraphics[width=100px,height=50px]{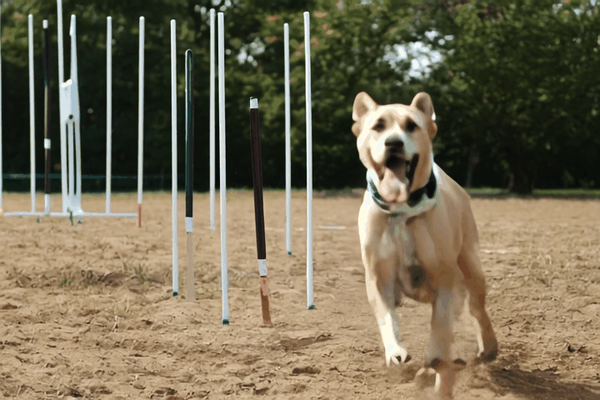} &
        \includegraphics[width=100px,height=50px]{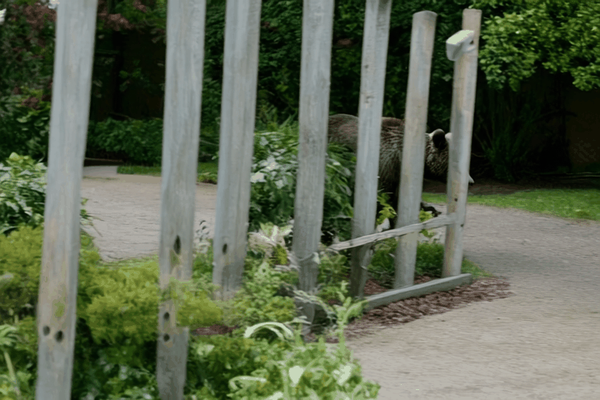} &
        \includegraphics[width=100px,height=50px]{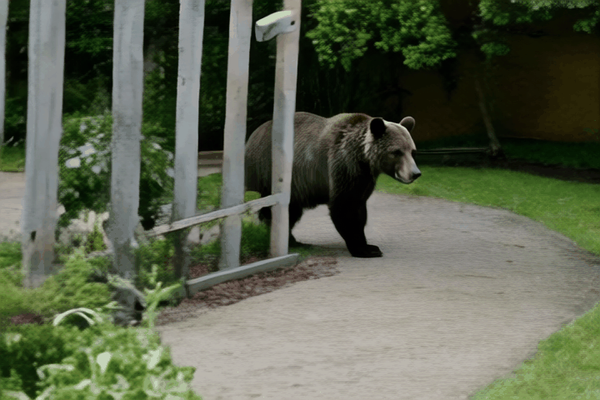} &
        \includegraphics[width=100px,height=50px]{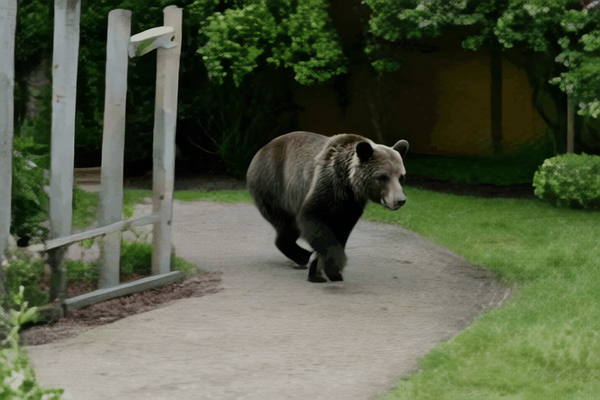} &
        \includegraphics[width=100px,height=50px]{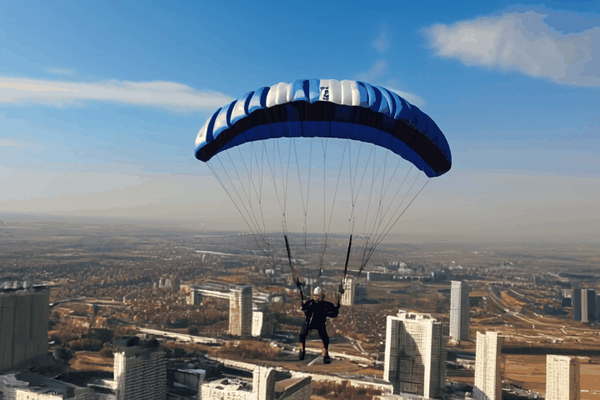} &
        \includegraphics[width=100px,height=50px]{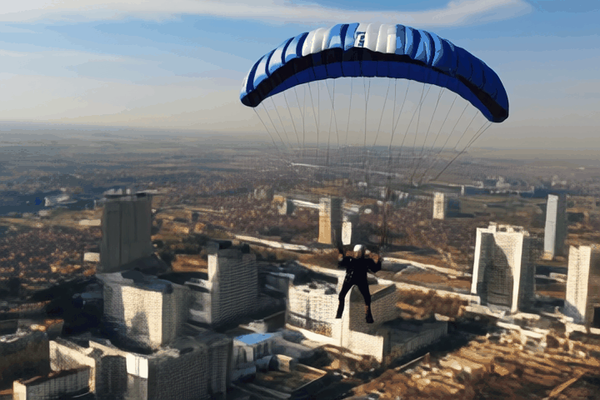} &
        \includegraphics[width=100px,height=50px]{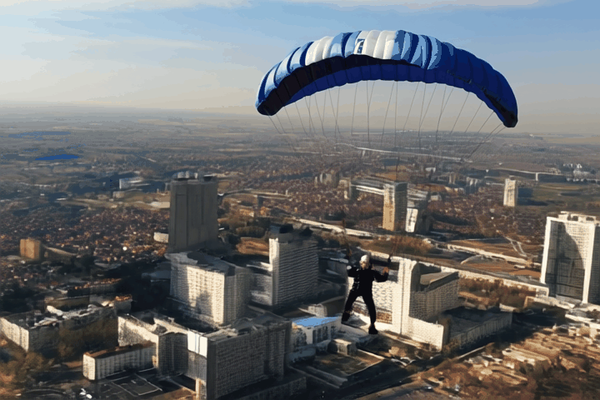} \\ %
        \multirow{1}{*}[39px]{\rotatebox{90}{MOFT~\cite{xiao2024video}}} & 
        \includegraphics[width=100px,height=50px]{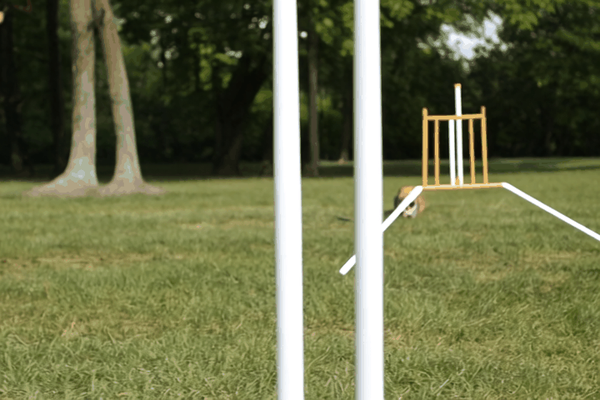} &
        \includegraphics[width=100px,height=50px]{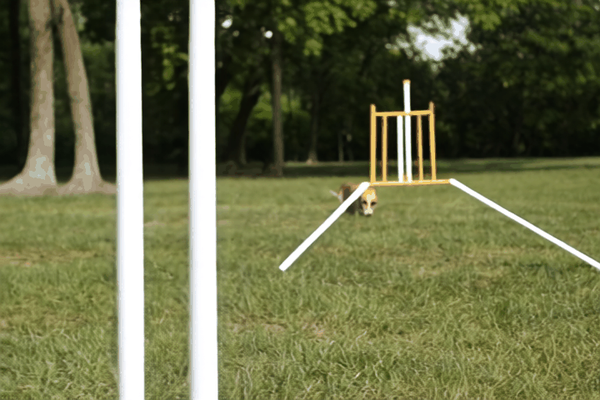} &
        \includegraphics[width=100px,height=50px]{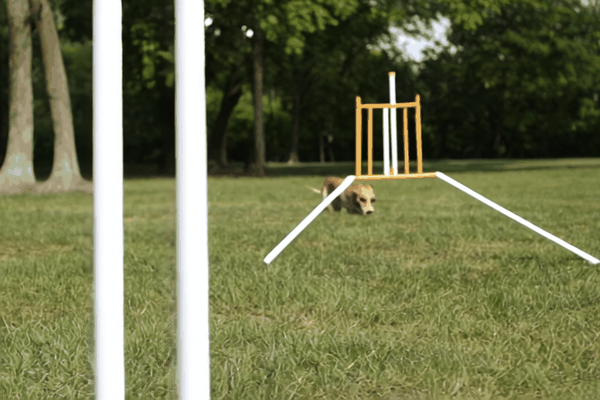} &
        \includegraphics[width=100px,height=50px]{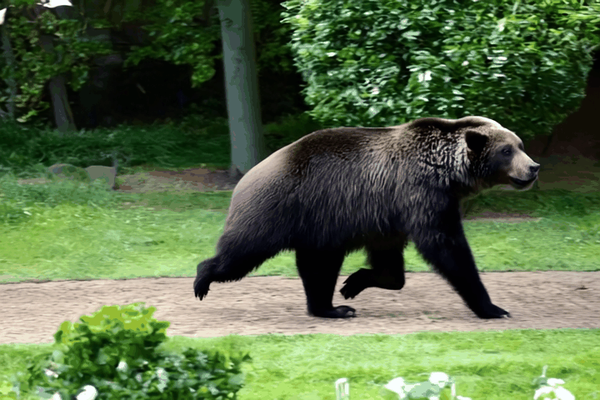} &
        \includegraphics[width=100px,height=50px]{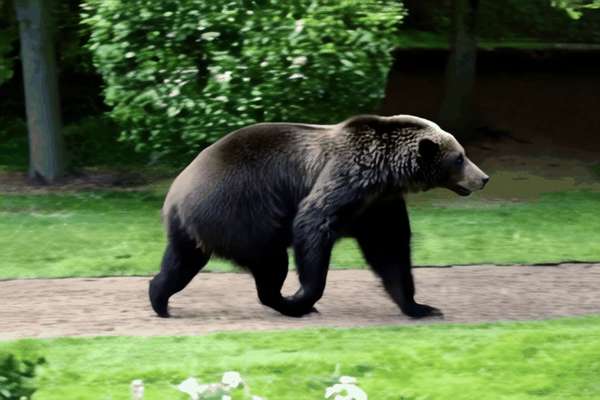} &
        \includegraphics[width=100px,height=50px]{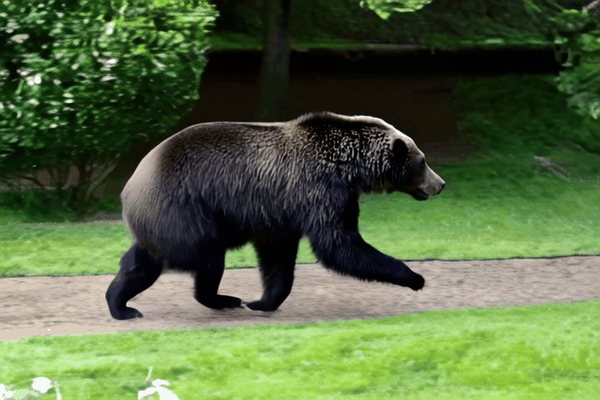} &
        \includegraphics[width=100px,height=50px]{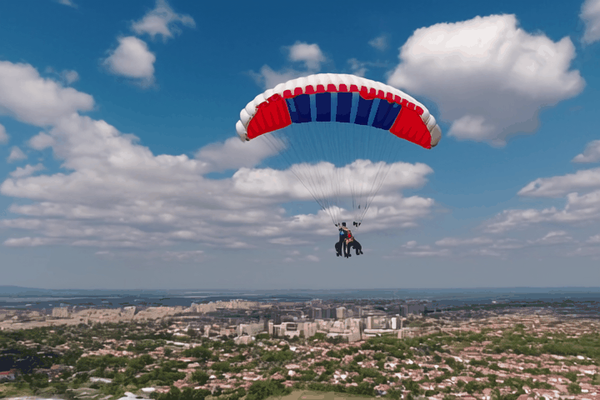} &
        \includegraphics[width=100px,height=50px]{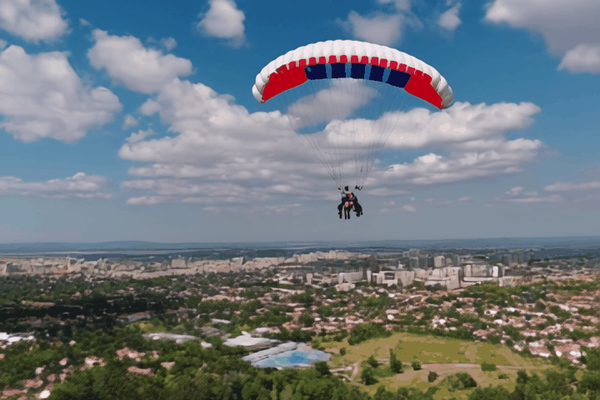} &
        \includegraphics[width=100px,height=50px]{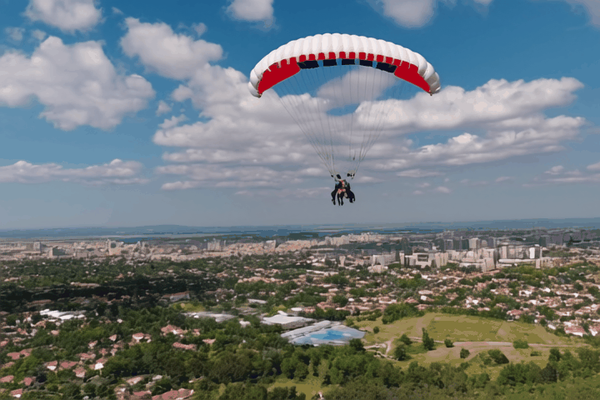} \\ %
        \multirow{1}{*}[39px]{\rotatebox{90}{\methodname}} & 
        \includegraphics[width=100px,height=50px]{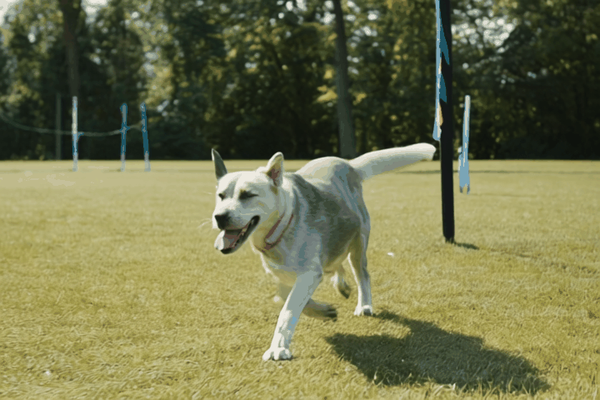} &
        \includegraphics[width=100px,height=50px]{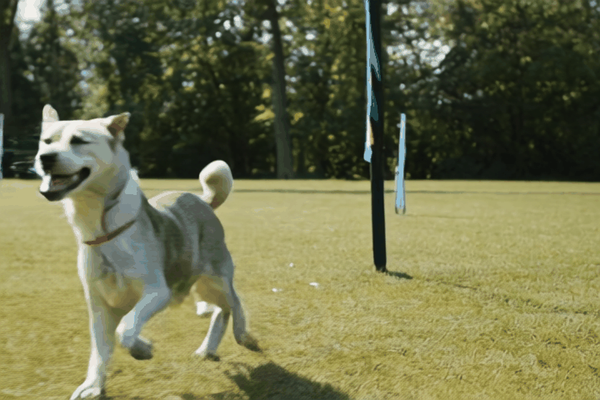} &
        \includegraphics[width=100px,height=50px]{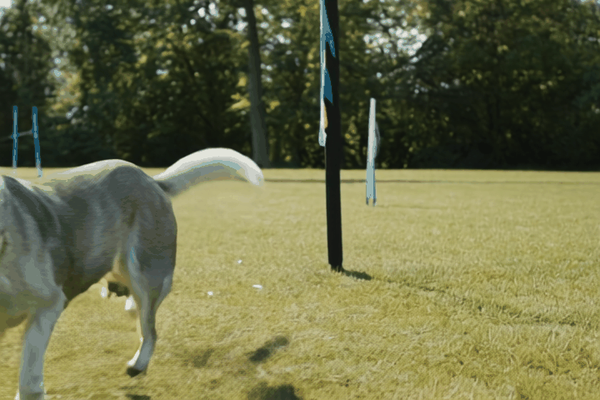} &
        \includegraphics[width=100px,height=50px]{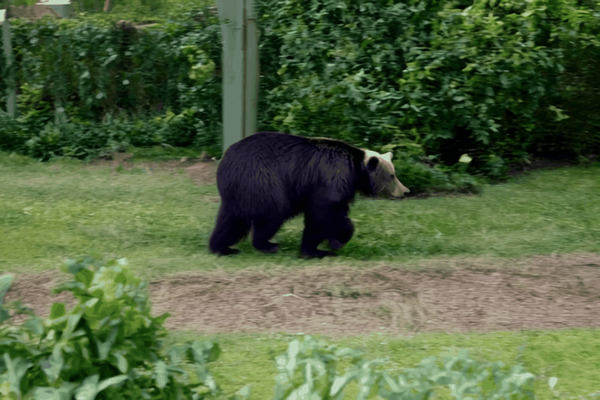} &
        \includegraphics[width=100px,height=50px]{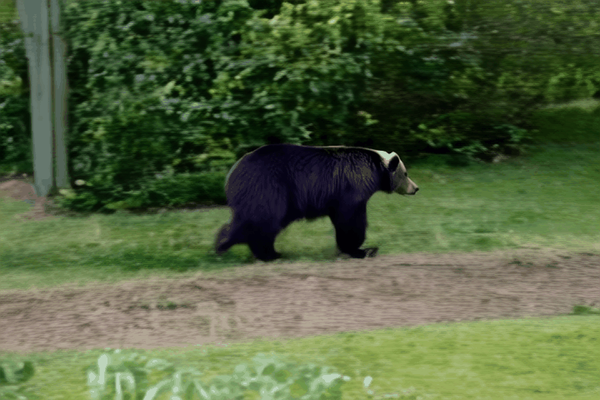} &
        \includegraphics[width=100px,height=50px]{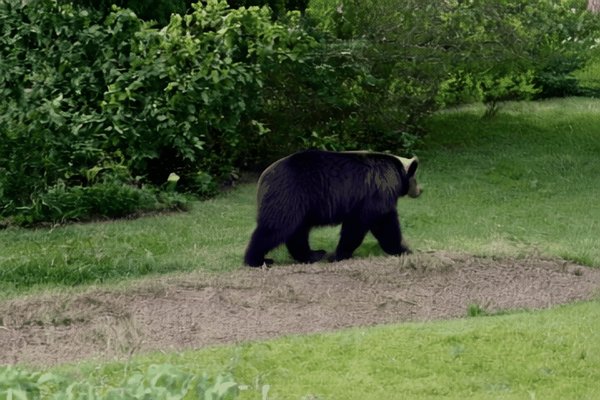} &
        \includegraphics[width=100px,height=50px]{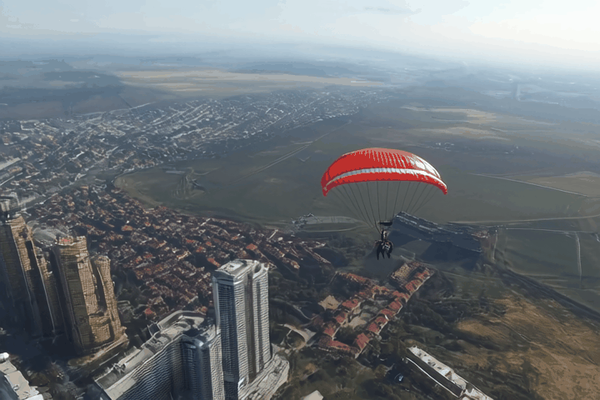} &
        \includegraphics[width=100px,height=50px]{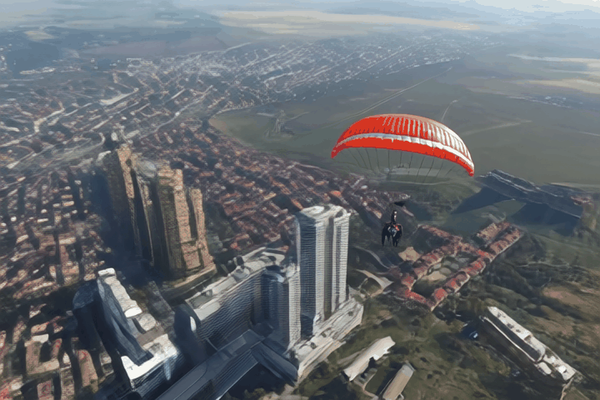} &
        \includegraphics[width=100px,height=50px]{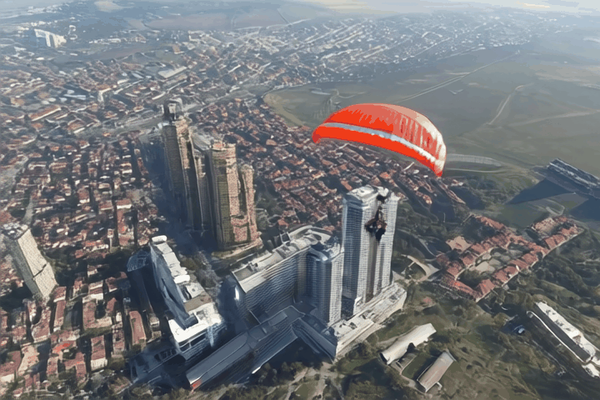} \\ %
        &\multicolumn{3}{c}{\hspace{-25px}``Dog running between poles in an agility course''} &         \multicolumn{3}{c}{\hspace{-25px}``Bear running in a garden''} &         \multicolumn{3}{c}{``Parachuting over a city, aerial view from above''} \\
        
    \end{tabular}}
    \caption{\textbf{Baseline comparison.} Baselines associate motion to wrong elements due to poor layout representation typical of UNet-based approaches that do spatial averaging or only consider deviations at each location. \methodname captures the spatio-temporal motion of each patch, resulting in correct spatial positioning and sizing of moving elements, \eg the dog (left), the bear (middle), the parachute (right).}
    \label{fig:qual}
\end{figure*}

\section{Experiments}\label{sec:exp}
\subsection{Experimental Setup}

\paragraph{Dataset and metrics.} 
For evaluation, we use 50 unique videos from the DAVIS dataset \cite{davis} coherently with the state-of-the-art~\cite{yatim2024space,xiao2024video}. To allow for a fine-grained motion transfer assessment, we test each video with three different prompts in order of similarity from the original video: (1) a \textit{\easy} prompt, created by simply captioning the video. This allows us to verify that the network disentangles the content from that of the original frames. (2) A \textit{\medium} prompt, obtained by changing the subject while keeping the background the same. (3) Lastly, a \textit{\hard} prompt, describing a completely different scene. This makes 150 motion-prompt pairs in total. For the evaluation, we use an \textit{Image Quality} (IQ) metric for frame-wise prompt fidelity assessment based on CLIPScore~\cite{hessel2021clipscore} and a \textit{Motion Fidelity} (MF) metric for motion tracklet consistency following~\cite{yatim2024space}. The MF metric~\cite{yatim2024space} compares the similarity of tracklets on $x$ and $x_\text{ref}$ obtained from off-the-shelf tracking~\cite{karaev2024cotracker}.

\paragraph{Baselines and networks.}
For video synthesis, we use the state-of-the-art DiT CogVideoX~\cite{yang2024cogvideox} with both 2 billion (CogVideoX-2B) and 5 billion (CogVideoX-5B) parameter variants. We compare against four baselines. First, we present a \textit{Backbone} method, which simply prompts the T2V model with $C$. This will act as a lower bound on MF. Following common practices~\cite{xiao2024video}, we define an \textit{Injection} baseline injecting extracted attention features during inference. Specifically, we inject keys $K$ and values $V$ obtained by processing $x_\text{ref}$ with the DiT at inference time. This loosely directs the structure of the synthesized scene, allowing it to roughly follow the spatial organization of elements in $x_\text{ref}$ without DDIM inversion~\cite{song2020denoising}. Given the simplicity and effectiveness of this technique, \textit{we apply it to all baselines} except for the Backbone. We evaluate against three optimization-based guidance methods: SMM~\cite{yatim2024space}, MOFT~\cite{xiao2024video} and MotionClone~\cite{motionclone}. We adapt them to CogVideoX for a fair comparison. For SMM, we replace the expensive DDIM inversion operation, taking over one hour per video on CogVideoX-5B, with KV injection. We adapt MotionClone for DiTs by operating on the temporal axis of attention. For fairness, we normalize the number of optimization steps. Due to the unavailability of the evaluation videos and prompts used in related works~\cite{yatim2024space,xiao2024video}, we test all baselines on our DAVIS setup.

\begin{figure*}[t]
    \centering
    \setlength{\tabcolsep}{2px}
    \resizebox{\linewidth}{!}{

    \begin{tabular}{cccc@{\hspace{20px}}ccc} %
        \multirow{1}{*}[34px]{\rotatebox{90}{Reference}} & 
        \includegraphics[width=100px,height=50px]{figures/bmx-trees_hard_climber/ref0.png} &
        \includegraphics[width=100px,height=50px]{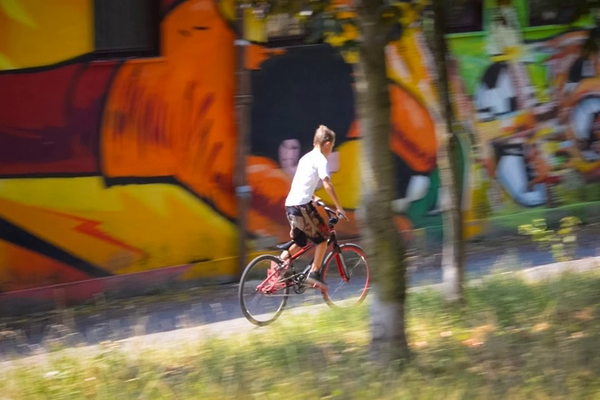} &
        \includegraphics[width=100px,height=50px]{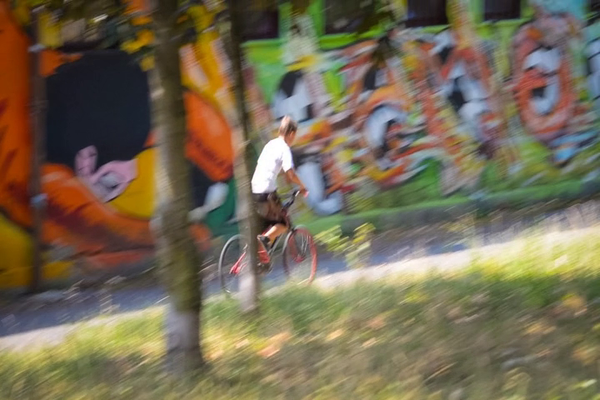} & 
        \includegraphics[width=100px, height=50px]{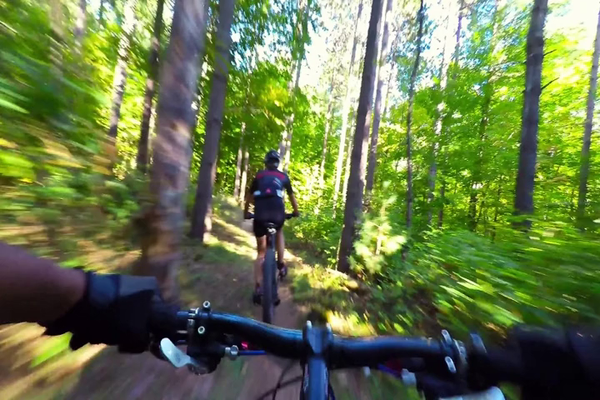} &
        \includegraphics[width=100px, height=50px]{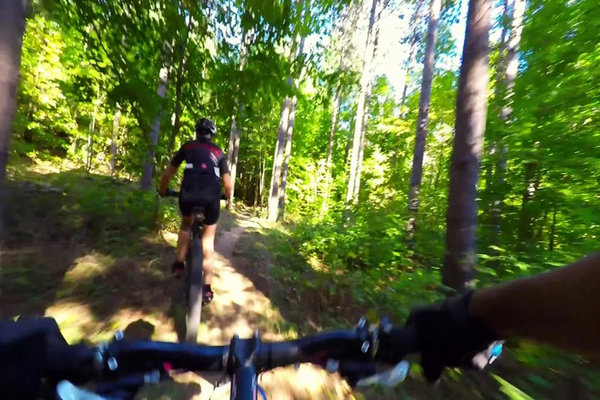} &
        \includegraphics[width=100px, height=50px]{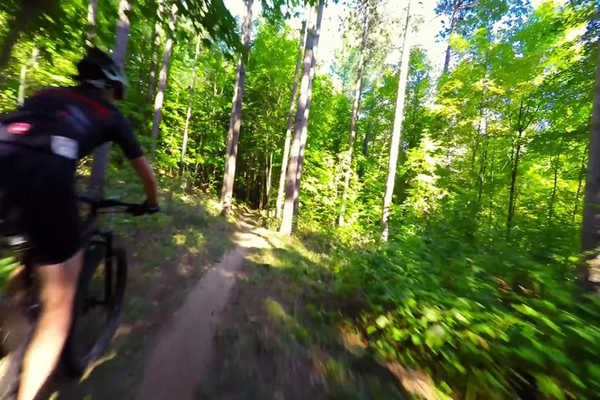} 
        
        \\
        \midrule
        & 
        \includegraphics[width=100px,height=50px]{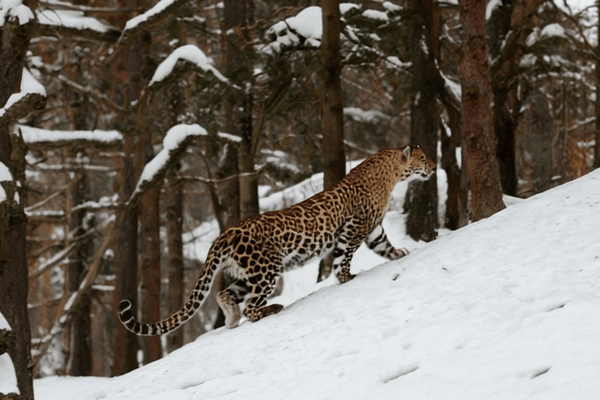} &
        \includegraphics[width=100px,height=50px]{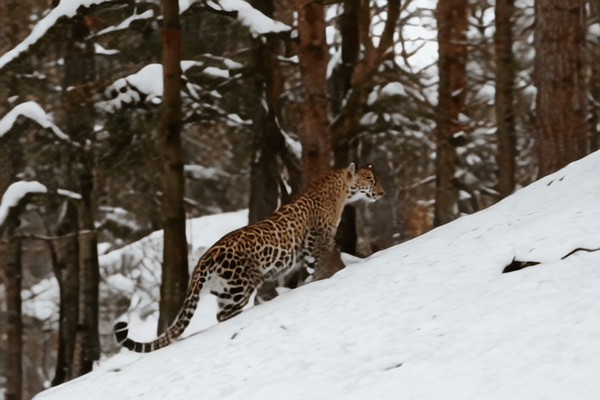} &
        \includegraphics[width=100px,height=50px]{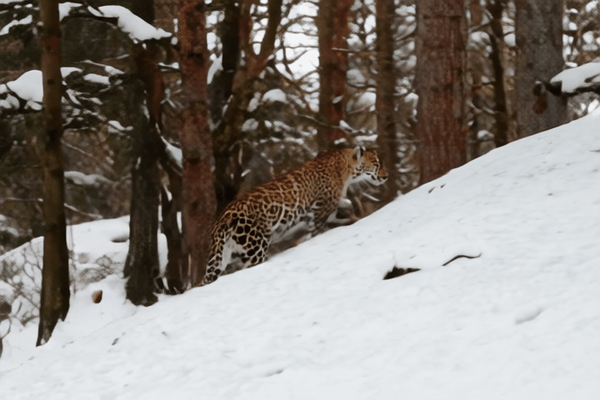}&
        \includegraphics[width=100px, height=50px]{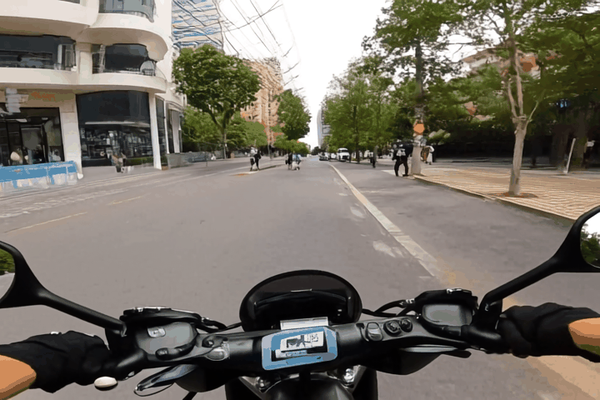} &
        \includegraphics[width=100px, height=50px]{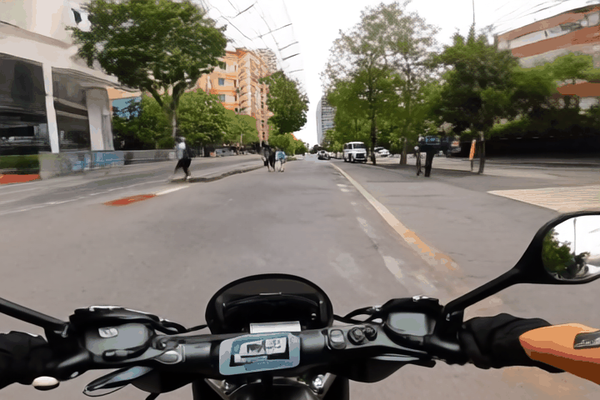} &
        \includegraphics[width=100px, height=50px]{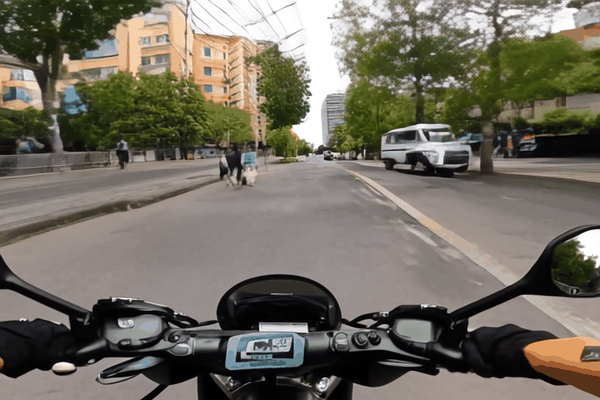}
        \\ 
        & \multicolumn{3}{c}{``Leopard running up a snowy hill in a forest''} & \multicolumn{3}{c}{``Driving motorcycle through cityscape, first person perspective''}\\
        \multirow{1}{*}[34px]{\rotatebox{90}{}} & 
        \includegraphics[width=100px,height=50px]{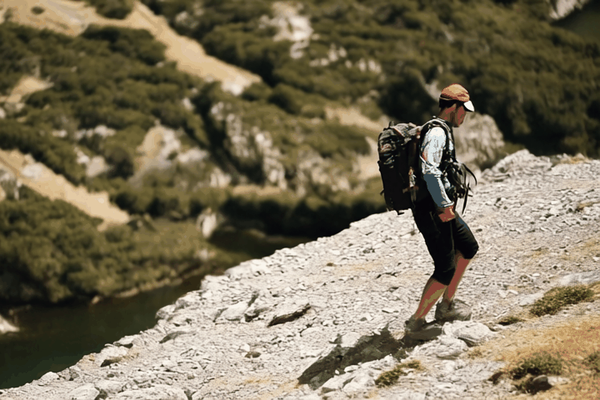} &
        \includegraphics[width=100px,height=50px]{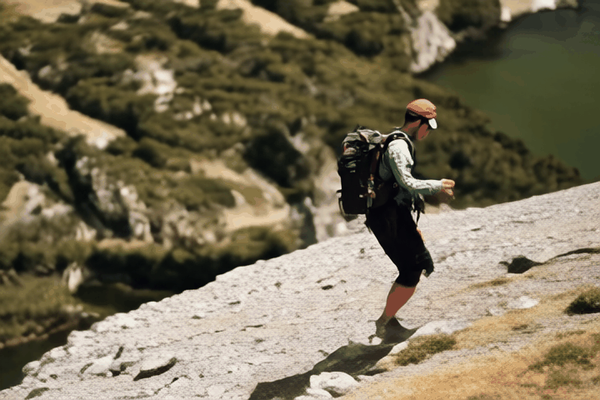} &
        \includegraphics[width=100px,height=50px]{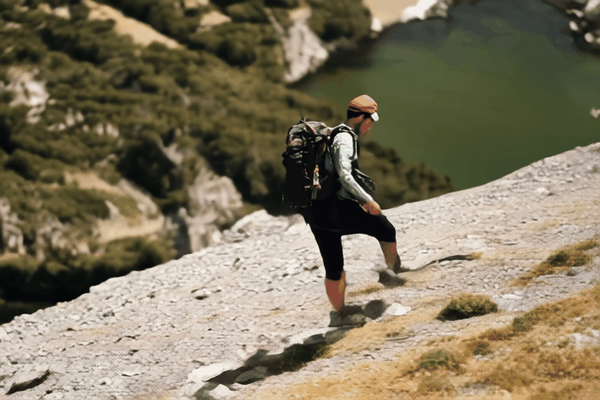}
        &\includegraphics[width=100px, height=50px]{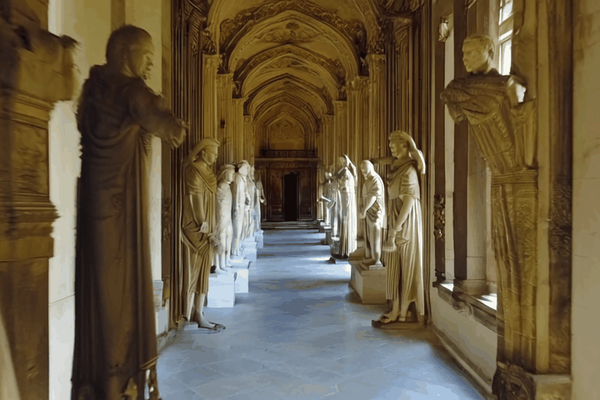} 
        &\includegraphics[width=100px, height=50px]{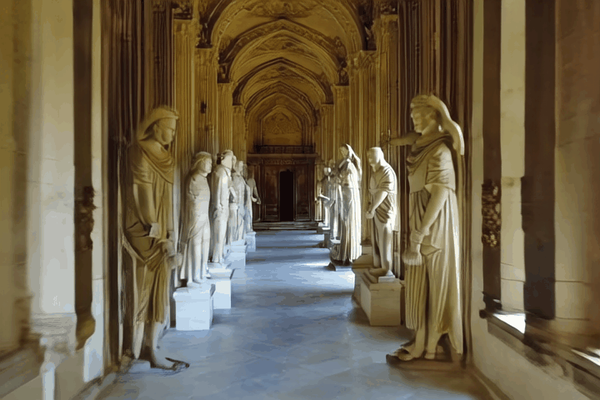} 
        &\includegraphics[width=100px, height=50px]{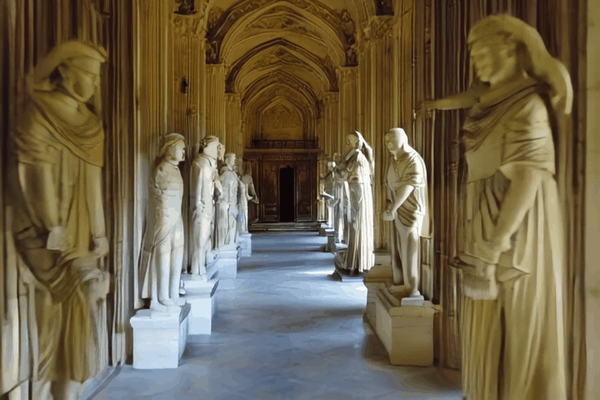}\\         
        & \multicolumn{3}{c}{``Hiker climbing upwards on a mountain peak''} & \multicolumn{3}{c}{``Drone footage of a castle corridor interior with tall statues''}\vspace{20px}\\
        \rotatebox{90}{Reference}
        & \includegraphics[width=100px, height=50px]{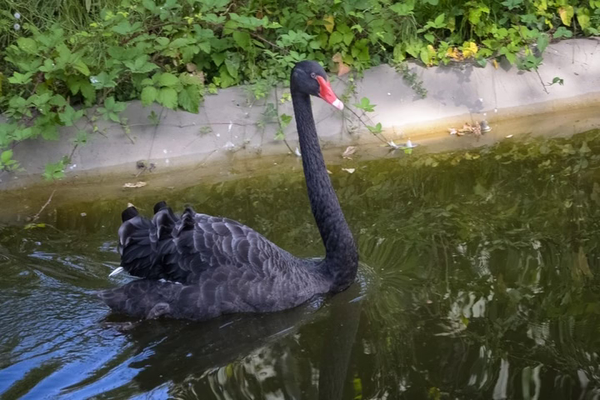} 
        &\includegraphics[width=100px, height=50px]{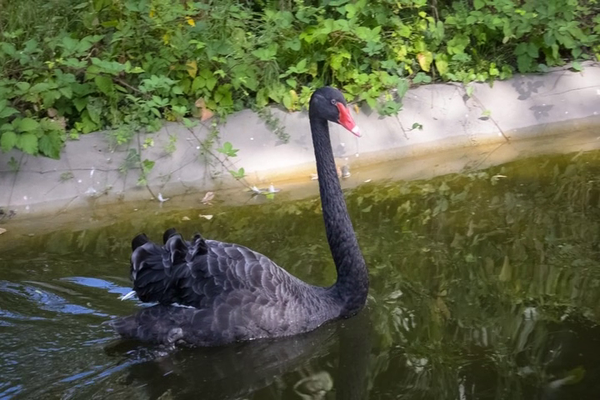} 
        &\includegraphics[width=100px, height=50px]{figures/blackswan_medium_boat/ref100.png} 
        & \includegraphics[width=100px, height=50px]{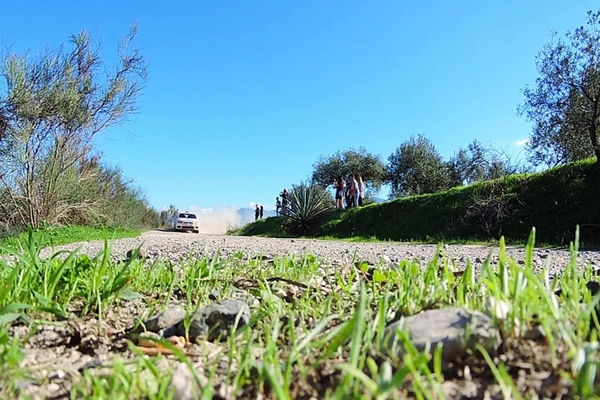} 
        &\includegraphics[width=100px, height=50px]{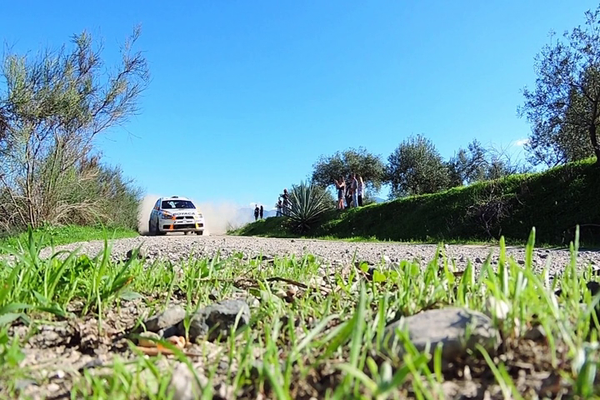} 
        &\includegraphics[width=100px, height=50px]{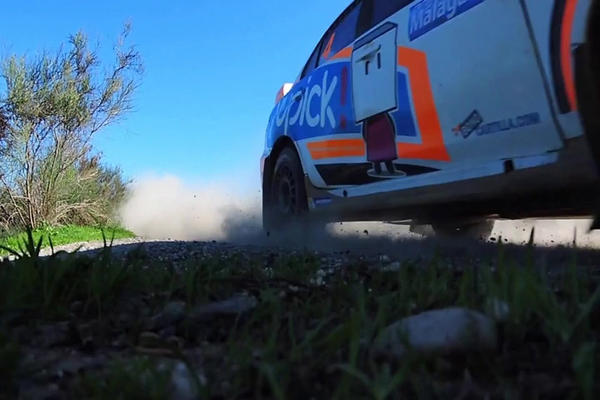}
        \\\midrule
        & \includegraphics[width=100px, height=50px]{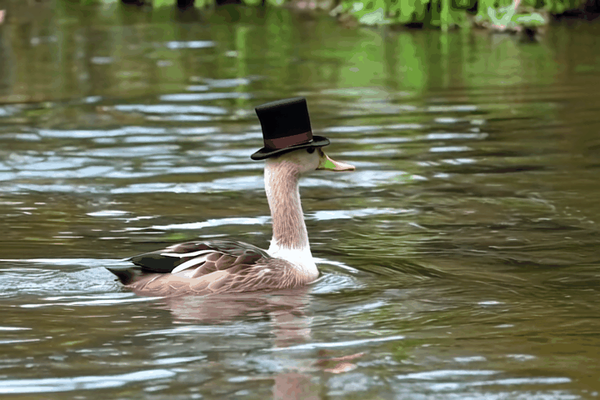} 
        &\includegraphics[width=100px, height=50px]{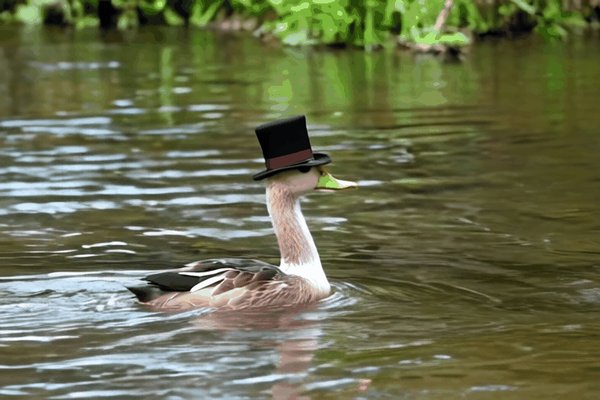} 
        &\includegraphics[width=100px, height=50px]{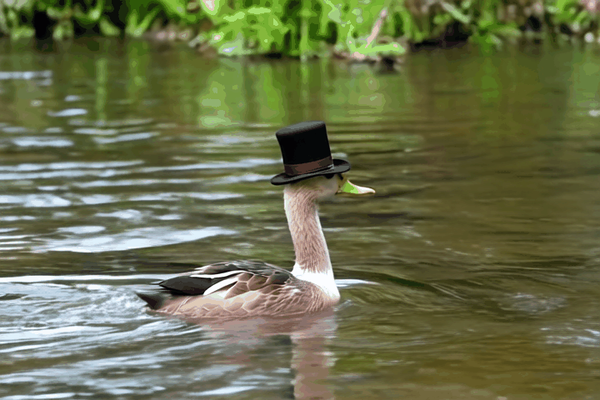} 
        & \includegraphics[width=100px, height=50px]{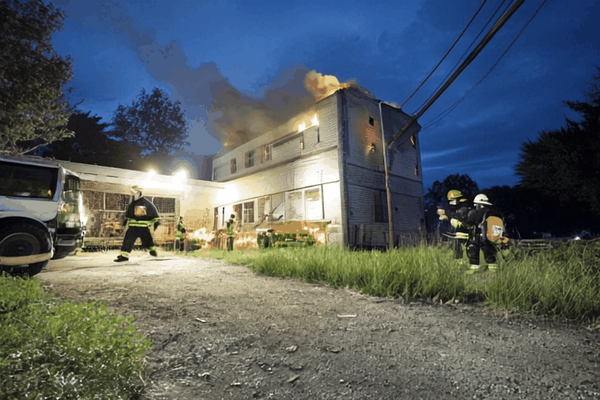} 
        &\includegraphics[width=100px, height=50px]{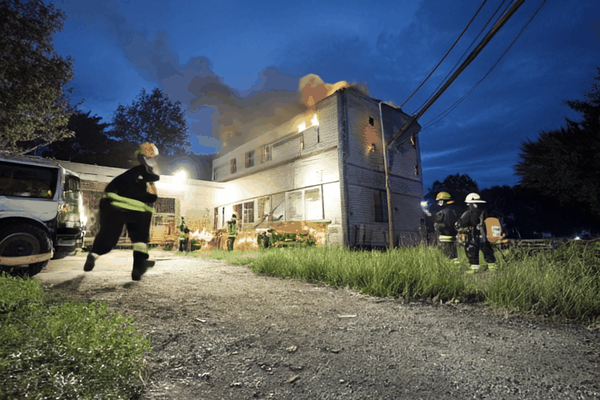} 
        &\includegraphics[width=100px, height=50px]{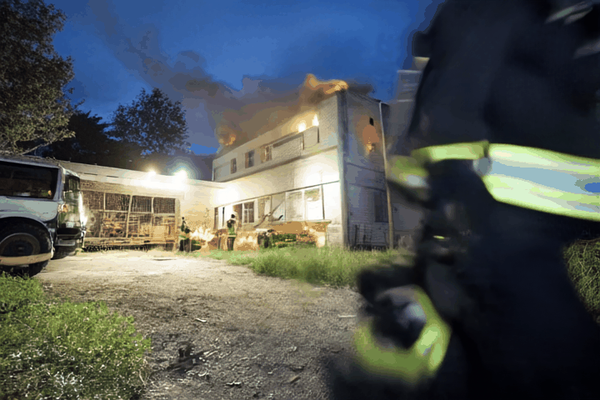}\\
        & \multicolumn{3}{c}{``A duck with a tophat swimming in a river''} & \multicolumn{3}{c}{``Firefighter running towards the camera away from a burning building''}\\
        & \includegraphics[width=100px, height=50px]{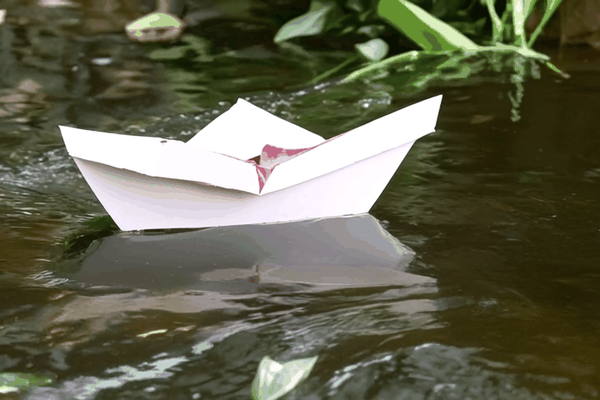} 
        &\includegraphics[width=100px, height=50px]{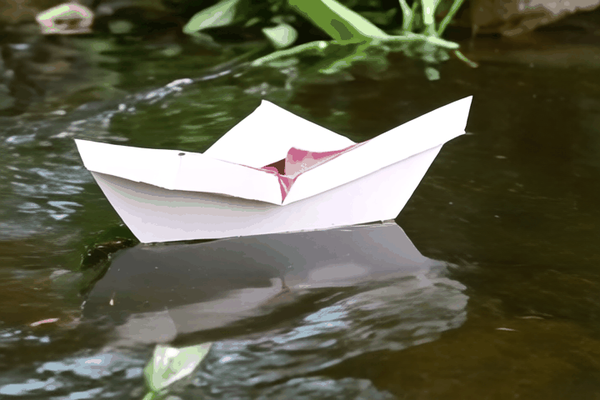} 
        &\includegraphics[width=100px, height=50px]{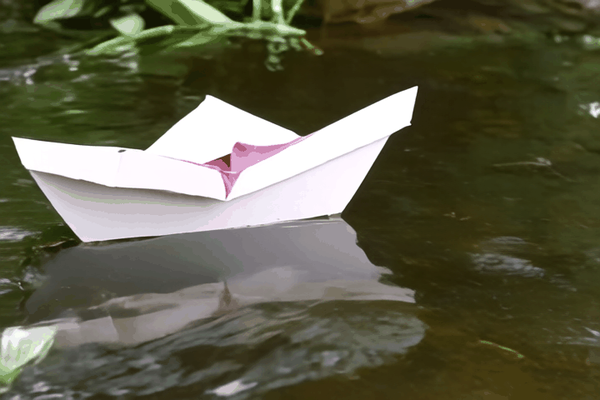} 
        & \includegraphics[width=100px, height=50px]{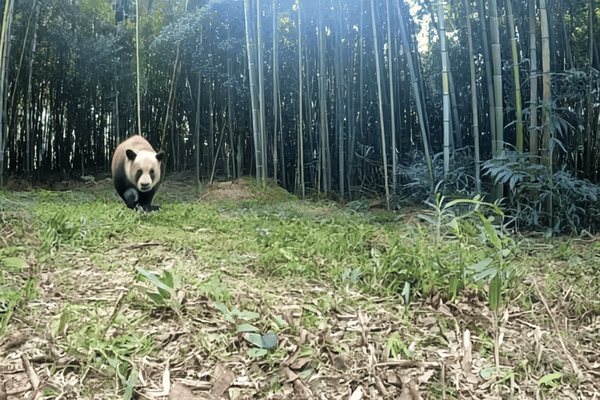} 
        &\includegraphics[width=100px, height=50px]{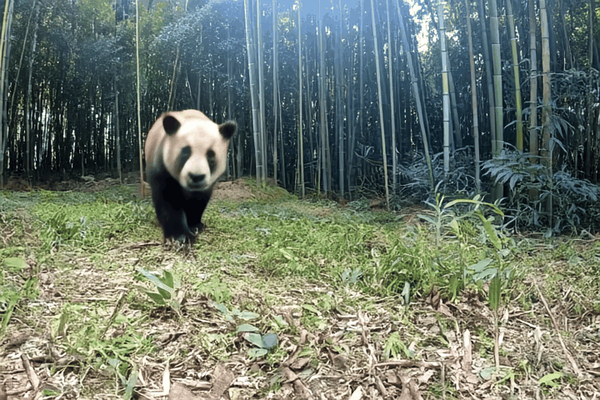} 
        &\includegraphics[width=100px, height=50px]{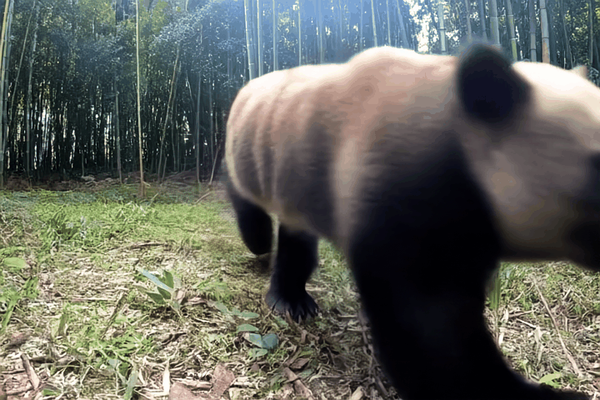}\\
        & \multicolumn{3}{c}{``A paper boat floating in a river''} & \multicolumn{3}{c}{``Panda charging towards the camera in a bamboo forest, low angle shot''}\\\\
        \end{tabular}}
    \caption{\textbf{Qualitative results of \methodname.} We are able to perform motion transfer in various conditions. Note how varying the prompt completely changes the scene's appearance while maintaining consistent motion. We map motion to correct elements even in cases where the motion changes drastically in positioning and size (bottom right).} 
    \label{fig:qual2}
\end{figure*}
\paragraph{Inference settings}
We employ 50 denoising steps for all baselines and optimize for 5 steps using Adam~\cite{adam} in the first 20\% of denoising timesteps with a linearly decreasing learning rate following~\cite{yatim2024space} from 0.002 to 0.001. We evaluate our AMF-based loss on the 20th block for CogVideoX-5B and the 15th for 2B. We apply KV injection on the first DiT block. Temperature $\tau$ is set to 2, emphasizing higher similarity tokens. On an NVIDIA A40 GPU, CogVideoX-2B generates a 21-frame video in around 3.5 minutes and 4 minutes with \methodname. CogVideoX-5B takes on average 5 minutes by default and 8 minutes with \methodname guidance. Unless explicitly specified, we optimize the latents $z_t$.

\subsection{Benchmarks}
\paragraph{Quantitative evaluation}
We evaluate the effectiveness of \methodname when optimizing $z_t$ in order to directly compare our AMF-based guidance with baselines. In the results presented in Table~\ref{tab:quant}, we consistently outperform all baselines. In particular, we considerably improve MF in both 5B and 2B models, scoring \textbf{0.785} and \textbf{0.726} respectively. In comparison, the best baseline, SMM, achieves \textbf{0.766} and \textbf{0.688}, demonstrating our superior capabilities to capture motion. Let us also highlight that when tested on \textit{\medium} prompts, SMM on CogVideoX-5B reports considerably lower MF (\textbf{0.741}) with respect to performance on \easy (\textbf{0.782}) and \hard (\textbf{0.776}). We attribute this to the entanglement of the guidance signal with $x_\text{ref}$. Notably, SMM is based on spatially-averaged global features extracted from the reference video. This makes it challenging to tackle semantic modifications impacting only part of the scene, such as \medium prompts. This is less evident for CogVideoX-2B due to the inferior overall performance. Conversely, \methodname and MOFT preserve local guidance, allowing for a more fine-grained semantic control on generated scenes. We outperform MotionClone as it does not take full advantage of the spatiotemporal information in attention. We observe that IQ values exhibit lower variability compared to setups with heterogeneous backbones~\cite{yatim2024space}. This shows that architecture is the main factor impacting image quality, further justifying our investigation of DiTs. 

\begin{figure}[t]
    \centering
    \includegraphics[width=\linewidth]{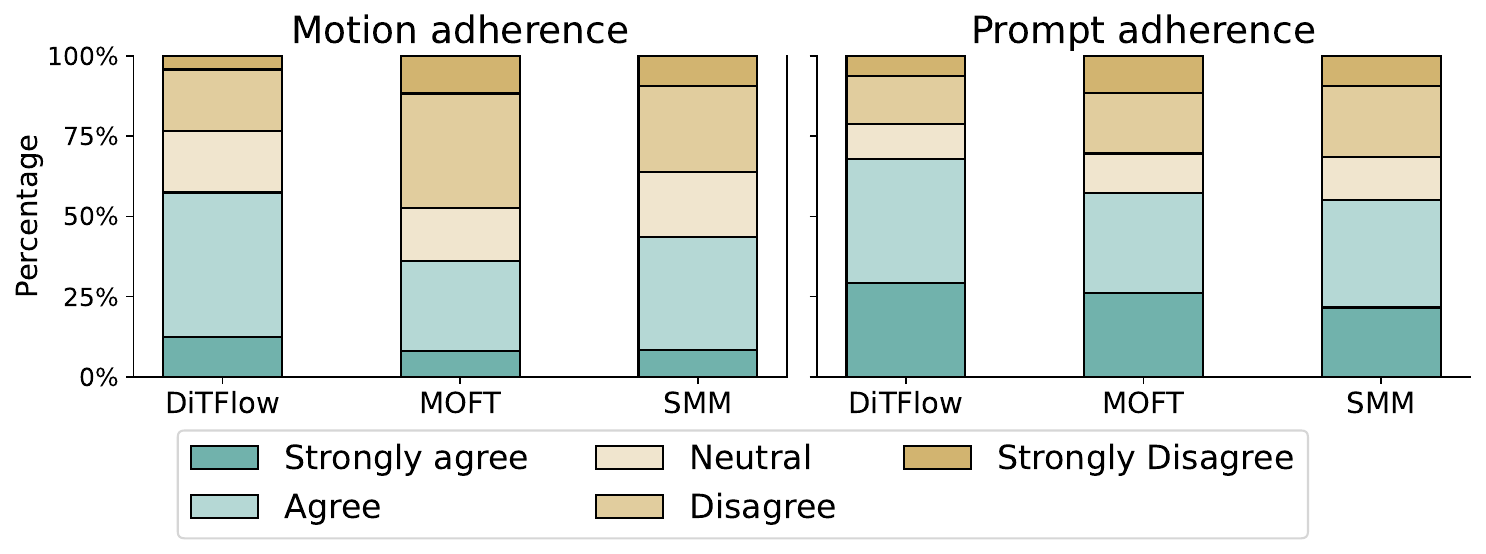}
    \caption{\textbf{Human evaluation.} We asked humans to evaluate agreement on the quality of generated samples in terms of motion (left) and prompt (right) adherence. \methodname consistently outperforms baselines in both evaluations.}
    \label{fig:user-study}
\end{figure}

\paragraph{Qualitative comparison.} We visually compare video generations of CogVideoX-5B against baselines in Figure~\ref{fig:qual}. MotionClone has less fine-grained disentanglement of motion as it directly guides attention values, while \textit{we guide the evolution of attentions over time}, regardless of their value. SMM suffers in \medium prompts, as observed quantitatively. SMM is based on a spatial averaging operation over features, which limits feature disentanglement as shown in the \medium column, where the rendered bars resemble those of the reference video. MOFT also tends to move the wrong elements in the scene as seen in the \easy example where the poles are moved instead of the dog. Their extracted MOFT feature guides the deviation of each spatial location from the temporal average, which can easily target the wrong content. \methodname, on the other hand, improves motion assignment by guiding the \textit{explicit relationship of patches across both space and time} through our AMF feature rather than guiding each location independently as in MOFT. Injection results are available in appendix.

We present additional qualitative results of \methodname in Figures~\ref{fig:qual2} and~\ref{fig:teaser} with realistic rendering of motion. We preserve motion in scenarios very different from the reference, such as in the top row. Motion is correctly mapped to specific elements in the scene even if they change drastically in size across frames, as in the challenging example in the bottom right. We attribute this to our patch-wise motion understanding, allowing it to capture fine-grained signals.

\paragraph{Human evaluation.}
We compare the best baselines on a study involving human judgment. In the first experiment (\textit{Motion Adherence}), we show users the reference video together with videos synthesized by \methodname and the best baselines on CogVideoX-5B. We ask in a Likert-5~\cite{likert1932technique} preference scale how much they agree with the statement ``The motion of the reference video is similar to the motion of the generated video''. Users were instructed to focus on motion dynamics and ignore content differences. In the second experiment, users assessed \textit{Prompt Adherence}, thus providing a video-level IQ score. They were asked to rate their agreement with the statement ``The text accurately describes the video''. We collect 66 preferences across 30 individuals, accounting for 1,980 unique answers. We report results in Figure~\ref{fig:user-study}. \methodname significantly outperforms both MOFT and SMM, consistent with the results in Table~\ref{tab:quant}.

\begin{figure}[t]
    \centering
    \begin{subfigure}{\linewidth}
        \begin{tikzpicture}
            \node[inner sep=0pt, outer sep=0pt, anchor=center] at (0,0) {\includegraphics[width=\linewidth]{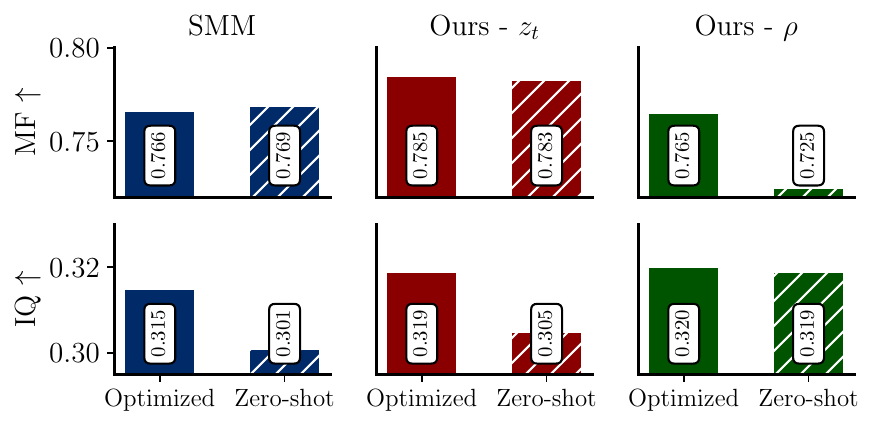}};
            \node[inner sep=0pt, outer sep=0pt, anchor=center] at (0,0.02) {\includegraphics[width=\linewidth]{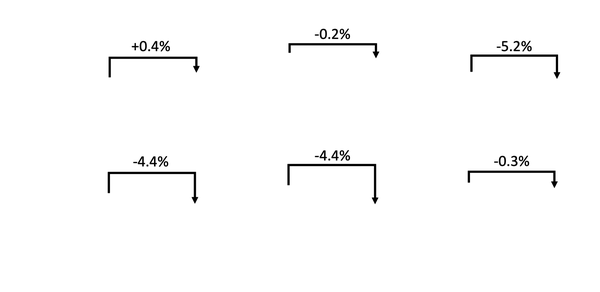}};
        \end{tikzpicture}
        \caption{Quantitative evaluation. SMM and Ours-$z_t$ follow Table~\ref{tab:quant} (All)}
        \label{fig:quant-zero-shot}
        \vspace{10px}

    \end{subfigure}
    \begin{subfigure}{\linewidth}
        \setlength{\tabcolsep}{1pt}
        \renewcommand{\arraystretch}{0.3}
        \resizebox{\linewidth}{!}{

            \begin{tabular}{ccc@{\hspace{10px}}cc}
                 
                  \multicolumn{5}{c}{\textbf{Prompts used for the optimization}}\\\midrule
                 &\multicolumn{2}{c}{\makecell{``Dog chasing geese in park''}} & \multicolumn{2}{c}{\makecell{``Zoom into a gorilla wearing \\ a lab coat on a field''}}\\
                 \midrule
                 \multirow{1}{*}[-13px]{\rotatebox{90}{SMM}}\\ & \includegraphics[width=60px,height=40px]{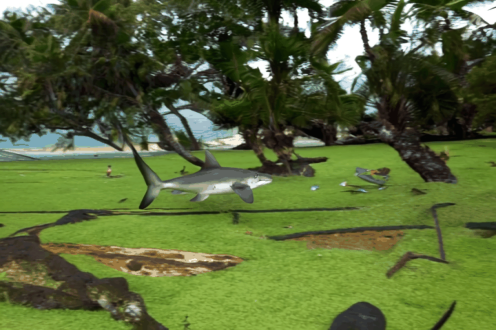} & \includegraphics[width=60px,height=40px]{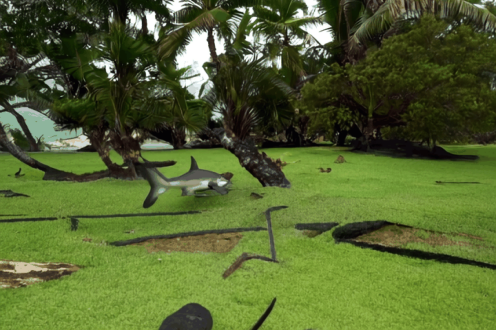} & \includegraphics[width=60px,height=40px]{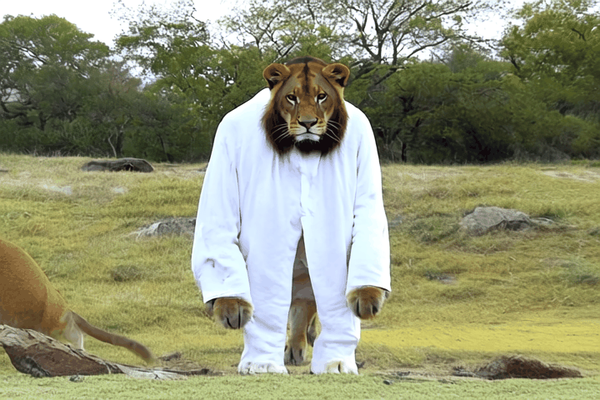} & \includegraphics[width=60px,height=40px]{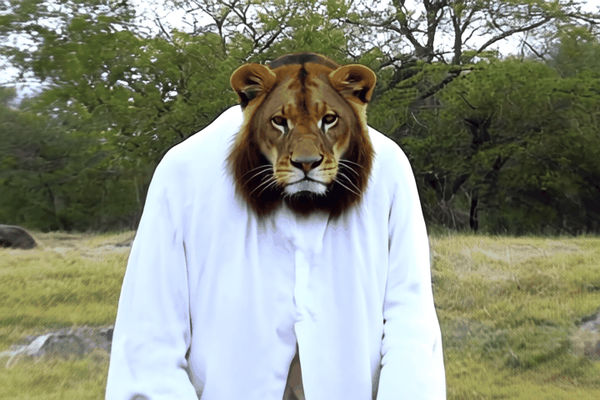} \\
                 \multirow{1}{*}[-8px]{\rotatebox{90}{Ours-$z_t$}}\\ & \includegraphics[width=60px,height=40px]{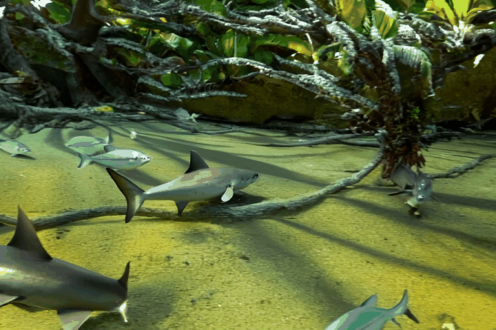} & \includegraphics[width=60px,height=40px]{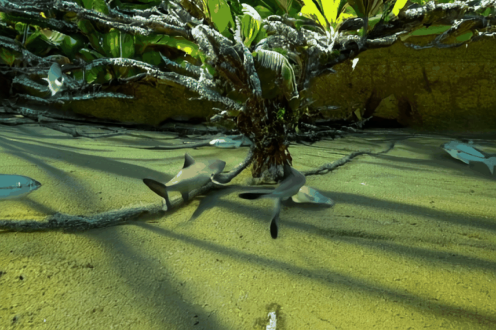} & \includegraphics[width=60px,height=40px]{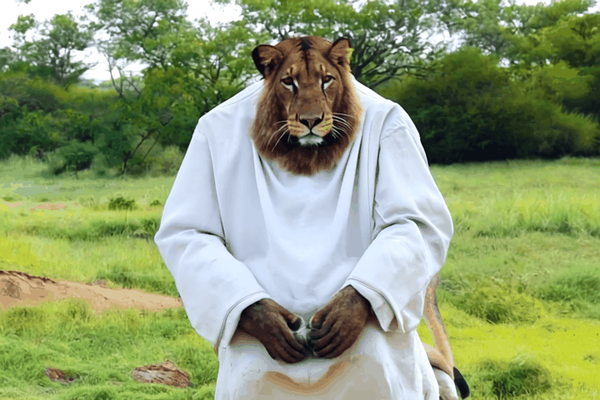} & \includegraphics[width=60px,height=40px]{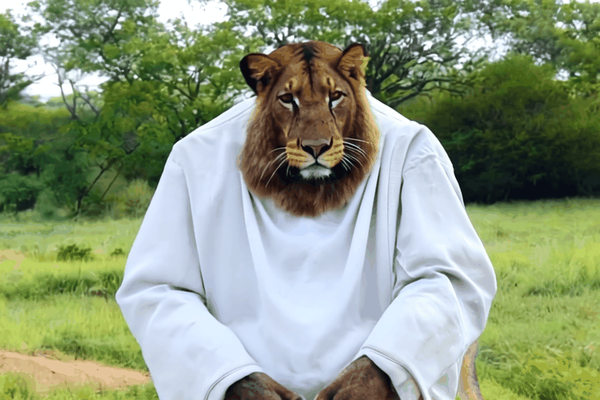} \\
                 \multirow{1}{*}[-8px]{\rotatebox{90}{Ours-$\rho$}}\\ & \includegraphics[width=60px,height=40px]{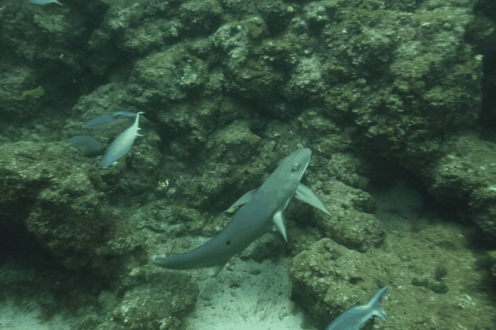} & \includegraphics[width=60px,height=40px]{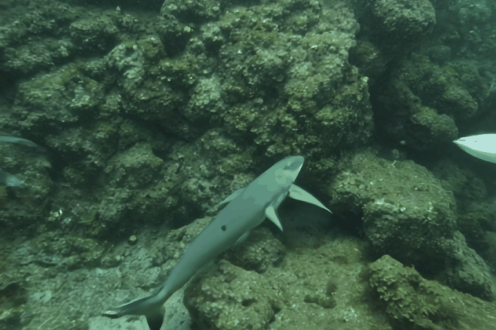} & \includegraphics[width=60px,height=40px]{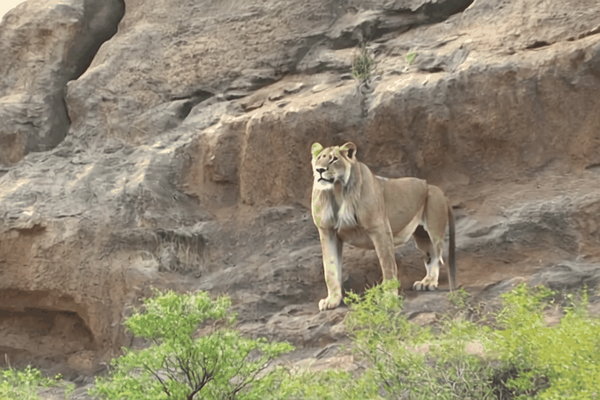} & \includegraphics[width=60px,height=40px]{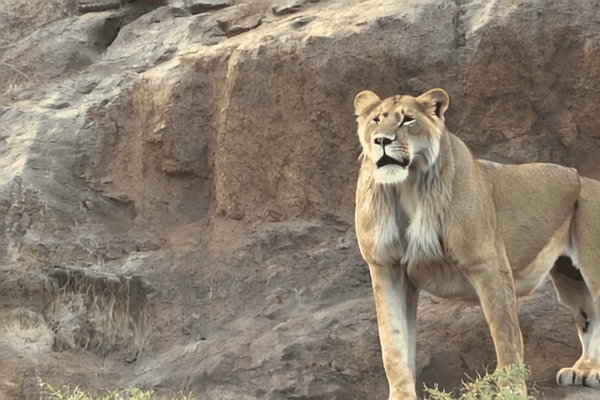} \vspace{5px}\\
                 &\multicolumn{2}{c}{\makecell{``Shark chasing fish in ocean''}} & \multicolumn{2}{c}{\makecell{``Zoom into a lion standing on a \\ cliff looking towards us''}}\\
            \end{tabular}
        }
    \caption{Qualitative evaluation}\label{fig:qual-zero-shot}

    \end{subfigure}
    \caption{\textbf{Zero-shot performance.} In~\subref{fig:quant-zero-shot}, we quantify zero-shot effectiveness. We compare performance by optimizing each prompt (Optimized) or using pre-optimized representations with new prompts (Zero-shot). Overall, optimizing $\rho$ allows for better preservation of IQ. This results in better zero-shot disentanglement when changing the prompt, as shown in~\subref{fig:qual-zero-shot}.}
\end{figure}

\subsection{Zero-shot generation}\label{sec:exp-zero-shot}
We evaluate the zero-shot capabilities of running \methodname with our optimized $\rho$ and a new prompt \textit{without further optimization}. We quantify performance across all prompt categories. We optimize $\rho$ on one prompt and infer with a different prompt. For instance, the optimized representation on the \easy prompt is injected into the generation of \medium and \hard prompts. Using the 5B model, we compare our methods to SMM~\cite{yatim2024space}, the best baseline method on MF in Table~\ref{tab:quant}. In Figure~\ref{fig:quant-zero-shot}, we report average zero-shot performance across all prompts. We define Ours-$z_t$ as \methodname with $z_t$ optimization, where the prompt is changed after optimizing $z_{0...t}$. Ours-$\rho$ is \methodname with positional embedding optimization following Section~\ref{sec:method-training}.

While optimizing $z_t$ leads to better zero-shot motion preservation (\textbf{-0.2\%}) compared to $\rho$ (\textbf{-5.2\%}), we report a significant drop in prompt adherence (\textbf{-4.4\%} vs \textbf{-0.3\%}) as seen in Figure~\ref{fig:qual-zero-shot}. Injecting the optimized $z_t$ partially leaks the content of the optimization prompt. For example, elements of the lab coat prompt appear in the lion generation, and park features carry over into the ocean setting on the left. Optimizing $\rho$ in the final row avoids this issue while rendering the desired motion, as the positional embedding has no impact on generated semantics, thus disentangling motion from content. Ultimately, with our findings on DiTs, we provide a novel adaptable strategy for prioritizing absolute performance through latent optimization or zero-shot capabilities using positional embeddings.

\begin{table}[t]
    \centering
    \setlength{\tabcolsep}{3px}
    \begin{subtable}{0.32\linewidth}
    \resizebox{\linewidth}{!}{
        \begin{tabular}{c|cc}
        \toprule
            \textbf{Block} & \textbf{MF} $\uparrow$ & \textbf{IQ} $\uparrow$ \\
            \midrule
            0 & 0.620 & 0.321\\
            10 & 0.670 & 0.309\\
            20 & \textbf{0.797} & \textbf{0.313}\\
            30 & 0.558 & 0.315\\

            \bottomrule            
        \end{tabular}
        }
        \caption{Guidance block}\label{tab:ablation-blocks}
    \end{subtable}
    \hfill
    \begin{subtable}{0.32\linewidth}
           \setlength{\tabcolsep}{3px}

        \resizebox{\linewidth}{!}{
    
        \begin{tabular}{c|cc}
        
        \toprule
            \textbf{$T_\text{opt}$} & \textbf{MF} $\uparrow$ & \textbf{IQ} $\uparrow$ \\
            \midrule
            0\% & 0.623 & 0.317\\
            20 \% & 0.797 & 0.313\\
            40 \% & \textbf{0.813} & \textbf{0.314}\\
            80 \% & 0.803 & 0.311\\
            100 \% & 0.799 & 0.312\\
            \bottomrule            
        \end{tabular}
        }
        \caption{Denoising steps}\label{tab:ablation-steps}

    \end{subtable}
    \hfill
    \begin{subtable}{0.32\linewidth}
        \resizebox{\linewidth}{!}{
    \setlength{\tabcolsep}{3px}
    \begin{tabular}{c|cc}
        \toprule
         \textbf{$K_\text{opt}$}&  \textbf{MF} $\uparrow$ & \textbf{IQ}$\uparrow$\\
         \midrule
         1 & 0.769 & 0.318 \\
         5 & 0.797 & 0.313 \\
         10 & \textbf{0.803} & \textbf{0.313} \\

         \bottomrule
    \end{tabular}
    }
    \caption{Optimization steps}\label{tab:opt-steps}
    \end{subtable}
    \caption{\textbf{Ablation studies.} We investigate our inference setups. In~\subref{tab:ablation-blocks}, we highlight that early blocks in DiTs contribute more to motion. In~\subref{tab:ablation-steps} and~\subref{tab:opt-steps}, we show that \methodname performance can be further boosted by increasing computational power.}
    \label{table:ablation}
\end{table}

\subsection{Ablation studies} \label{sec:ablation}
We ablate our design choices for \methodname on CogVideoX-5B. In Table~\ref{table:ablation}, we show the impact of 1) the DiT block index where we optimize, 2) the denoising iterations with guidance and 3) the number of optimization steps by varying each in isolation. We run \methodname on 14 unique videos and corresponding prompts. In Table~\ref{tab:ablation-blocks}, we show that the first blocks of CogVideoX-5B have incremental importance in motion guidance. We select the 20th block yielding best metrics. We notice that increasing the number of denoising (Table~\ref{tab:ablation-steps}) and optimization steps (Table~\ref{tab:opt-steps}) positively impacts motion transfer at the cost of more computation. In particular, we report best performance for 40\% of $T$ (MF \textbf{0.813}) and 10 optimization steps (MF 0.803). Each of these setups double the optimization time for each generation, hence we selected 20\% of $T$ and 5 optimization steps for the best speed/quality tradeoff. Nevertheless, this enables improved motion transfer by investing in computation.

%% file: sec/6_conclusions.tex
\section{Conclusions}\label{sec:conclusions}

We present \methodname, the first DiT-specific method for motion transfer, based on our novel AMF formulation. Through extensive experiments and multiple evaluations, we demonstrate the effectiveness of \methodname against baselines in motion transfer, with extensions to zero-shot generation. Improved video action representations in DiTs may lower costs for generating robotic simulations and enable intuitive control of subject actions in content creation.

%% file: sec/7_acknowledgments.tex
\section{Acknowledgments}\label{sec:acknowledgments}

The authors extend their thanks to Constantin Venhoff and Runjia Li for their helpful feedback on the first draft. We appreciate the time put in by the participants of our user study. Alexander Pondaven is generously supported by a Snap studentship and the EPSRC Centre for Doctoral Training in Autonomous Intelligent Machines and Systems [EP/S024050/1]. We would also like to thank Ivan Johnson and Adam Lugmayer for their endless support with the compute server.

%% file: sec/X_suppl.tex
\clearpage
\setcounter{page}{1}
\setcounter{section}{0}
\maketitlesupplementary
\appendix

We strongly encourage readers to check the qualitative video samples in the project page at \url{ditflow.github.io}. Here, we provide additional elements for easing the understanding of our work. Specifically, we first provide implementation details in Section~\ref{sec-supp:implementation} and further ablations in Section~\ref{sec-supp:moreablation}. Then, we provide additional reasoning about alternative strategies for supervision (Section~\ref{sec-supp:ablations}) and propose a simple experiment to justify AMF in Section~\ref{sec-supp:amfjustify}. Further MotionClone evaluation is provided in Section~\ref{sec-supp:mclone}. Finally, we discuss limitations (Section~\ref{sec-supp:limitations}).

\section{Implementation}\label{sec-supp:implementation}

\paragraph{Positional embedding training details} CogVideoX-5B uses a different positional embedding mechanism to CogVideoX-2B. CogVideoX-2B uses 3D sinusoidal embeddings similar to \cite{transformer} and these are simply added to the tokens to provide absolute positional information. During guidance, gradients can backpropagate from the AMF loss at block 15 to these embeddings. CogVideoX-5B uses 3D rotary positional embeddings (RoPE \cite{rope}) that are embedded into all queries and keys at each attention block. Gradients still backpropagate from block 20 to the RoPE applied to all previous blocks.

\paragraph{Dataset}
We provide a sample of the dataset in Table~\ref{tab:dataset}. Video names are the same as those used in the DAVIS dataset \cite{davis}. Please refer to the supplementary material included in the project page for the full dataset and visuals.

\section{Additional Ablation Studies}\label{sec-supp:moreablation}

We conducted further ablation studies on temperature ($\tau$), number of guidance blocks, and optimization algorithm in Table~\ref{tab:further-ablation}, following the experimental setup in Section~\ref{sec:ablation}. The temperature ablation reveals that setting $\tau=5$ yields a marginal improvement in both MF and IQ metrics. Importantly, the performance varies only slightly across different temperature values, demonstrating DiTFlow's robustness to this hyperparameter. We also compare our original single-block approach (using only block 20) against a multi-block configuration (blocks 20+15+10). While DiTFlow benefits slightly from multi-block guidance, we opted for the single-block approach in our main experiments due to the additional computational overhead associated with multi-block configurations. Finally, we compare results for three different optimizers and find that Adam has the best balance between motion transfer and quality.

\begin{table}[t]
\centering
\setlength{\tabcolsep}{3px}
\begin{subtable}{0.32\linewidth}
\resizebox{\linewidth}{!}{
\begin{tabular}{c|cc}
\toprule
\textbf{Temperature} & \textbf{MF} $\uparrow$ & \textbf{IQ} $\uparrow$ \\
\midrule
1 & 0.763 & 0.313 \\
2 & 0.797 & 0.313\\
5 & \textbf{0.799} & \textbf{0.317} \\
10 & 0.777 & 0.315 \\
\bottomrule
\end{tabular}
}
\caption{Temperature}\label{tab:ablation-temperature}
\end{subtable}
\hfill
\begin{subtable}{0.32\linewidth}
\setlength{\tabcolsep}{3px}
\resizebox{\linewidth}{!}{
    \begin{tabular}{c|cc}
    \toprule
        \textbf{Blocks} & \textbf{MF} $\uparrow$ & \textbf{IQ} $\uparrow$ \\
        \midrule
        20 & 0.797 & \textbf{0.313} \\
        10,15,20 & \textbf{0.804} & \textbf{0.313} \\
        \bottomrule
    \end{tabular}
    }
    \caption{Multi-block setup}\label{tab:ablation-multiblock}
\end{subtable}
\hfill
\begin{subtable}{0.32\linewidth}
    \resizebox{\linewidth}{!}{
\setlength{\tabcolsep}{3px}
\begin{tabular}{c|cc}
    \toprule
     \textbf{Optimizer}&  \textbf{MF} $\uparrow$ & \textbf{IQ}$\uparrow$\\
     \midrule
     Adam & 0.797 & 0.313 \\
     AdamW & \textbf{0.803} & 0.311 \\
     SGD & 0.623 & \textbf{0.320} \\
     \bottomrule
\end{tabular}
}
\caption{Optimizer algorithm}\label{tab:ablation-optimizer}
\end{subtable}
\caption{\textbf{Further ablations on CogVideoX-5B.}}
\label{tab:further-ablation}
\end{table}

\section{Nearest neighbor alternatives}\label{sec-supp:ablations}

\begin{figure}[t]
    \centering
    \begin{subfigure}{0.45\linewidth}
        \centering
        \includegraphics[width=\linewidth]{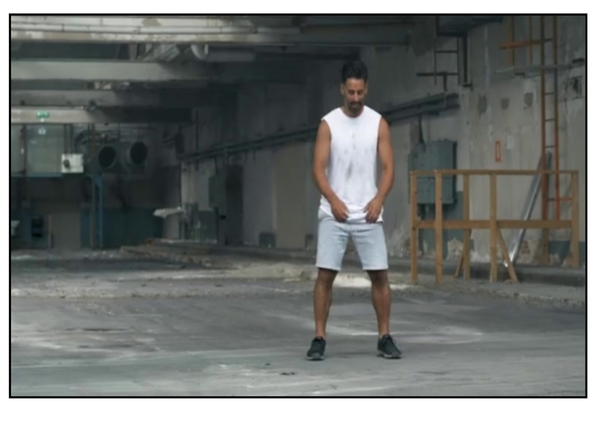}
        \caption{Frame $i$}\label{tab:supp_framei}
    \end{subfigure}
    \hfill
    \begin{subfigure}{0.45\linewidth}
        \centering
        \includegraphics[width=\linewidth]{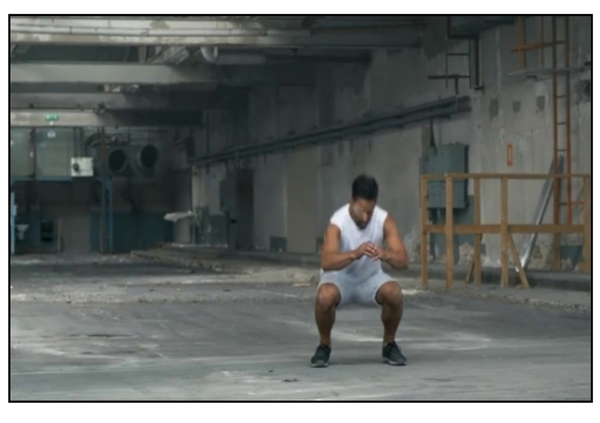}
        \caption{Frame $j$}\label{tab:supp_framej}
    \end{subfigure}

    \begin{subfigure}{0.45\linewidth}
        \centering
        \includegraphics[width=\linewidth]{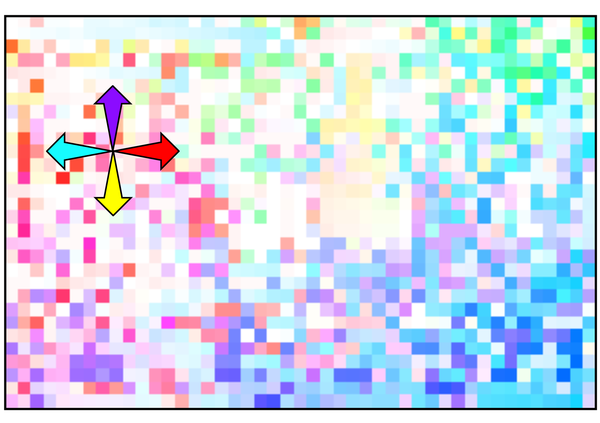}
        \caption{Latent nearest neighbour}\label{tab:supp_latentnn}
    \end{subfigure}
    \hfill
    \begin{subfigure}{0.45\linewidth}
        \centering
        \includegraphics[width=\linewidth]{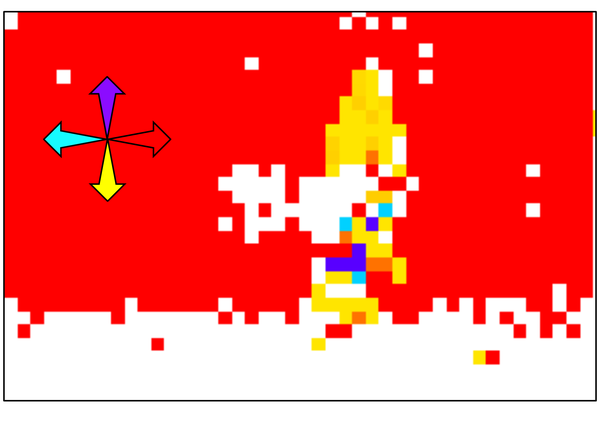}
        \caption{AMF displacement $\Delta_{i,j}$}\label{tab:supp_argmax}
    \end{subfigure}
    
    \caption{\textbf{Displacement maps of squat motion.} We visualise the displacement map between frames ~\subref{tab:supp_framei} and ~\subref{tab:supp_framej} computed on latents. The displacements are mapped to colours according to the colour wheel arrows shown. Taking the latent nearest neighbour~\cite{geyer2023tokenflow} in ~\subref{tab:supp_latentnn} results in very noisy displacements with poor matching of content between frames. The AMF displacement in ~\subref{tab:supp_argmax} captures the downwards (yellow) motion of the person and rightwards (red) motion of the panning camera better.}
    \label{fig:amf_visual}
\end{figure}

An alternative signal for AMF construction could have been the usage of nearest neighbors on noisy latents, as in related works~\cite{geyer2023tokenflow}. In Figure~\ref{fig:amf_visual}, we visualize correspondences extracted between two frames using this technique and compare it to our AMF displacement. We demonstrate a much smoother displacement map, which can lead to better guidance on the rendered video. 

\section{Justification of AMF experiment}\label{sec-supp:amfjustify}

We conduct a small-scale study on 14 videos (from Section~\ref{sec:ablation}), where we move the content of each video's first frame in a random direction. We calculate the AMF of each video. If the motion of content is correctly captured, the AMF vectors should point in the direction of the introduced motion vector. We calculate the patch-wise cosine similarity between AMF and ground truth motion. We obtain 0.857 for CogVideoX-2B and 0.734 for CogVideoX-5B. The lower bound is 0.5 if random directions are predicted. This proves that \textit{AMF is a valid signal for capturing motion}, which also aligns with the AMF visualisation in Figure~\ref{fig:amf_visual}. The superior performance of the 5B model can also be attributed to its better motion representations.

\section{Injection baseline}\label{sec-supp:mclone}
\begin{figure*}[t]
    \centering
    \setlength{\tabcolsep}{2px}
    \resizebox{\linewidth}{!}{

    \begin{tabular}{ccHc@{\hspace{20px}}cHc@{\hspace{20px}}cHc} %
        &\multicolumn{3}{c}{\hspace{-25px}\textbf{\easy}} &         \multicolumn{3}{c}{\hspace{-25px}\textbf{\medium}} &         \multicolumn{3}{c}{\textbf{\hard}} \\
        \multirow{1}{*}[34px]{\rotatebox{90}{Reference}} & 
        \includegraphics[width=100px,height=50px]{figures/dog-agility_easy/ref1.png} &
        \includegraphics[width=100px,height=50px]{figures/dog-agility_easy/ref2.png} &
        \includegraphics[width=100px,height=50px]{figures/dog-agility_easy/ref3.png} &
        \includegraphics[width=100px,height=50px]{figures/libby_medium/ref1.png} &
        \includegraphics[width=100px,height=50px]{figures/libby_medium/ref2.png} &
        \includegraphics[width=100px,height=50px]{figures/libby_medium/ref3.png} &
        \includegraphics[width=100px,height=50px]{figures/paragliding_hard/ref1.png} &
        \includegraphics[width=100px,height=50px]{figures/paragliding_hard/ref2.png} &
        \includegraphics[width=100px,height=50px]{figures/paragliding_hard/ref3.png} \\ %
        \midrule
        \multirow{1}{*}[34px]{\rotatebox{90}{Injection}} & 
        \includegraphics[width=100px,height=50px]{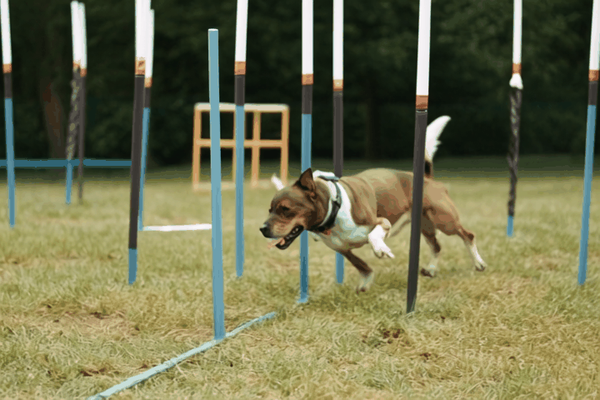} &
        \includegraphics[width=100px,height=50px]{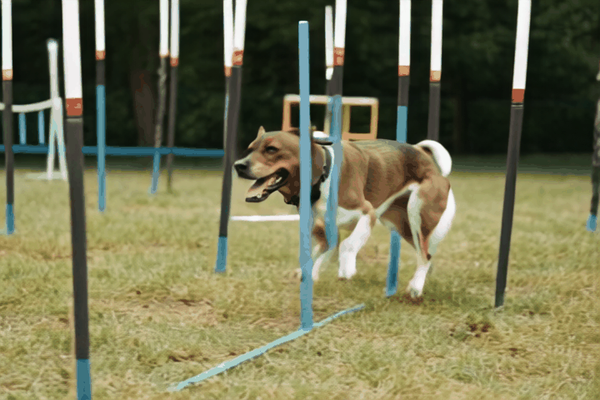} &
        \includegraphics[width=100px,height=50px]{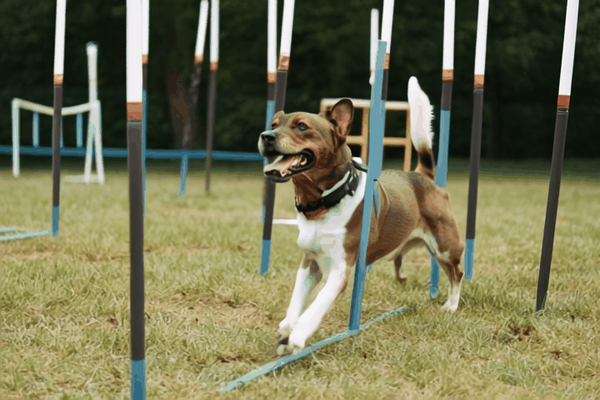} &
        \includegraphics[width=100px,height=50px]{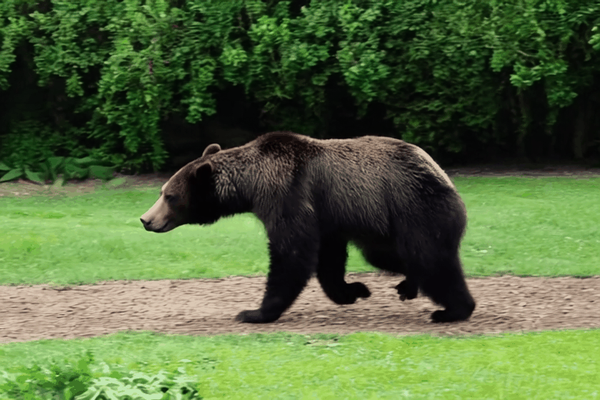} &
        \includegraphics[width=100px,height=50px]{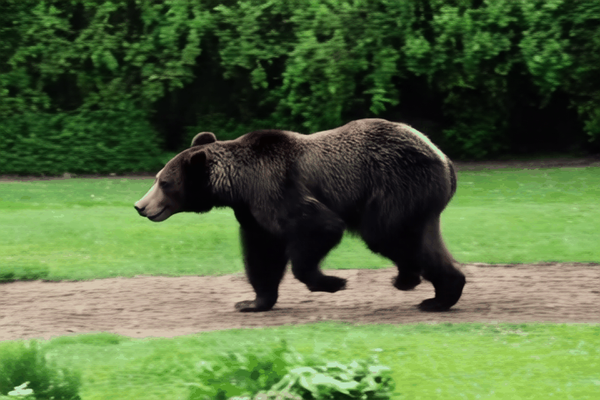} &
        \includegraphics[width=100px,height=50px]{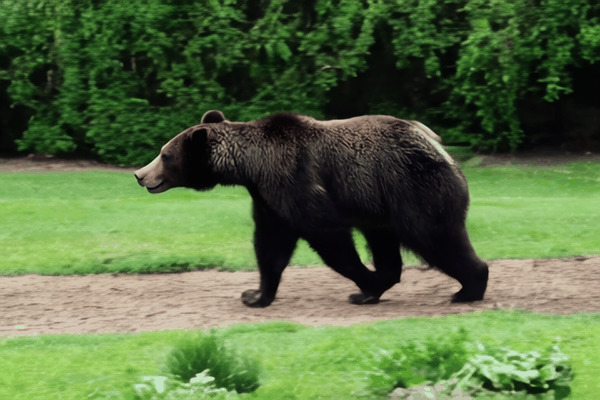} &
        \includegraphics[width=100px,height=50px]{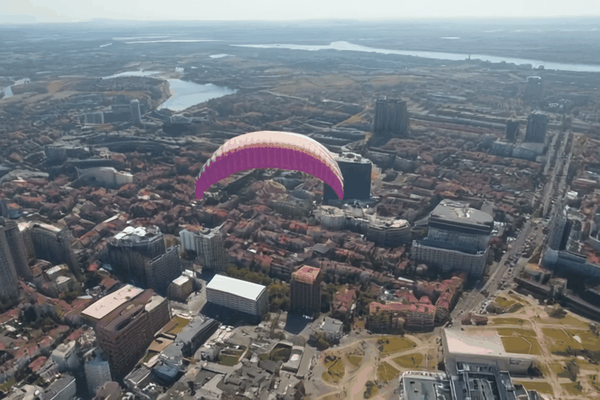} &
        \includegraphics[width=100px,height=50px]{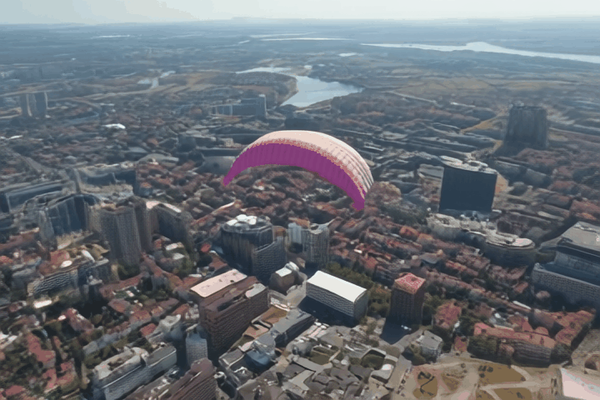} &
        \includegraphics[width=100px,height=50px]{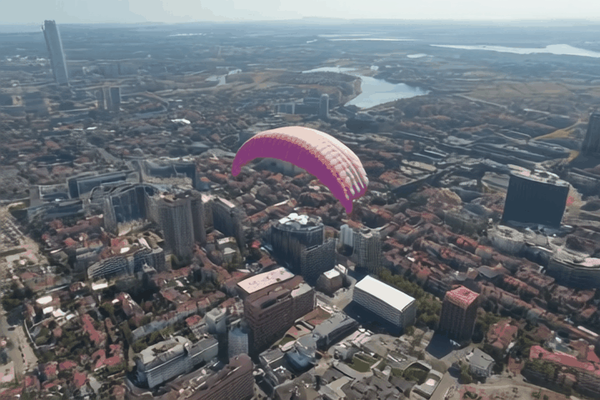} \\ %
        \multirow{1}{*}[35px]{\rotatebox{90}{\methodname}} & 
        \includegraphics[width=100px,height=50px]{figures/dog-agility_easy/flowloss_latentopt1.png} &
        \includegraphics[width=100px,height=50px]{figures/dog-agility_easy/flowloss_latentopt2.png} &
        \includegraphics[width=100px,height=50px]{figures/dog-agility_easy/flowloss_latentopt3.png} &
        \includegraphics[width=100px,height=50px]{figures/libby_medium/flowloss_latentopt1.png} &
        \includegraphics[width=100px,height=50px]{figures/libby_medium/flowloss_latentopt2.png} &
        \includegraphics[width=100px,height=50px]{figures/libby_medium/flowloss_latentopt3.png} &
        \includegraphics[width=100px,height=50px]{figures/paragliding_hard/flowloss_latentopt1.png} &
        \includegraphics[width=100px,height=50px]{figures/paragliding_hard/flowloss_latentopt2.png} &
        \includegraphics[width=100px,height=50px]{figures/paragliding_hard/flowloss_latentopt3.png} \\ %
        &\multicolumn{3}{c}{\hspace{-25px}``Dog running between poles in an agility course''} &         \multicolumn{3}{c}{\hspace{-25px}``Bear running in a garden''} &         \multicolumn{3}{c}{``Parachuting over a city, aerial view from above''} \\
        
    \end{tabular}}
    \caption{\textbf{Injection baseline.} Results are provided in the same setup as Figure~\ref{fig:qual}.}
    \label{fig:motionclonequal}
\end{figure*}

We provide the qualitative results of our Injection baseline in Figure~\ref{fig:motionclonequal}. It is able to transfer coarse subject location information while deviating significantly from the reference motion. For instance, in the \medium example, the bear walks in the opposite direction.

\section{Limitations}\label{sec-supp:limitations}
As seen in previous methods \cite{yatim2024space}, generations are still limited to the pre-trained video generator, so it has difficulty transferring motion with prompts or motions that are out of distribution. For example, complex body movements (\eg \textit{backflips}) still remain a difficult task for these models. Moreover, we highlight that motion transfer is inherently ambiguous if not associated to prompts. For example, transferring the motion of a dog to a plane may risk to map motion features of other elements in the scene to the plane in the rendered video, even with KV injection. For future work, we believe it will be important to associate specific semantic directions (e.g. $\text{dog}\mapsto\text{plane}$) to constrain editing, similar to what happens in inversion-based editing~\cite{mokady2022nulltext}.

The pairwise nature of AMF does lead to slightly more memory consumption compared to previous methods. However, we found performance to be consistent across setups with different number of frames. Long video generation approaches that generate a smaller set of frames at a time may readily be applied to our motion transfer approach.

\begin{table*}[t]
   \centering
   \begin{tabular}{|l|p{0.27\textwidth}|p{0.27\textwidth}|p{0.27\textwidth}|}
   \hline
   \textbf{Video} & \textbf{Caption} & \textbf{Subject} & \textbf{Scene} \\
   \hline
   blackswan & A black swan swimming in a river & A duck with a tophat swimming in a river & A paper boat floating in a bathtub \\
   \hline
   car-turn & A gray car with black tires driving on a road in a forest & A man with a black top running on a road in a forest, camera shot from a distance & Black suv with tinted windows driving through a roundabout in a bustling city, surrounded by tall buildings and bright lights \\
   \hline 
   car-roundabout & A gray mini cooper driving around a roundabout in a town & A man riding a unicycle around a roundabout in a town & A lion walking through a bustling roundabout, surrounded by vibrant city life \\
   \hline
   libby & Dog running in a garden & Bear running in a garden & Plane flying through the sky above the clouds \\
   \hline
   bus & Aerial view of bus driving on a street & Aerial view of red ferrari driving on a street & Closeup aerial view of an ant crawling in a desert \\
   \hline
   camel & A camel walking in a zoo & A giraffe walking in a zoo & A blue Sedan car turning into a driveway \\
   \hline 
   bear & A bear walking on the rocks & A giraffe walking on the rocks & A giraffe walking in the zoo \\
   \hline
   bmx-bumps & BMX rider biking up a sandy hill & Black Jeep driving up a sandy hill & Black Jeep driving up a hill in a bustling city \\
   \hline
   bmx-trees & Kid with white shirt riding a bike up a hill, seen from afar, long-distance view & Leopard running up a grassy hill & Leopard running up a snowy hill in a forest \\
   \hline
   boat & Fishing boat sails through the sea in front of an island, close-up, medium shot, elevated camera angle, wide angle view & Black yacht sails through the sea in front of an island & Black yacht sails through the sea in front of a bustling city \\
   \hline
   \end{tabular}
   \caption{\textbf{Dataset snippet.} Sample of DAVIS videos chosen with associated prompts from each category described in Section~\ref{sec:exp}.}
   \label{tab:dataset}
\end{table*}